%% file: main.tex
\documentclass[journal]{IEEEtran}
\IEEEoverridecommandlockouts

\usepackage{cite}
\usepackage{graphicx}
\usepackage[utf8]{inputenc}
\usepackage[normalem]{ulem}

\usepackage{booktabs, multicol, multirow}
\usepackage[table]{xcolor}
\usepackage{tabularx}

\if CLASSINFOpdf
\else

\fi

\usepackage{amsmath}
\usepackage{hyperref}
\usepackage{cleveref}
\usepackage[linesnumbered,algoruled,boxed,lined]{algorithm2e}
\usepackage{amsfonts}
\usepackage{algorithmic}
\usepackage{graphicx}
\usepackage{float}  
\usepackage{subcaption}
\usepackage{xcolor}

\usepackage[switch]{lineno}
\renewcommand{\color}[1]{}

\begin{document}
\title{EndoL2H: Deep Super-Resolution for Capsule Endoscopy}
%
%
%


\author{
        Yasin~Almalioglu,
        Kutsev~Bengisu~Ozyoruk,
        Abdulkadir~Gokce,
        Kagan~Incetan,
        Guliz Irem Gokceler
        Muhammed~Ali~Simsek,
        Kivanc Ararat,
        Richard J. Chen,
        Nicholas J. Durr, 
        Faisal~Mahmood,
        Mehmet~Turan
        \thanks{(Corresponding Author: Mehmet Turan. This work was supported by the Scientific and Technological Research Council of Turkey (TUBITAK) for 2232 - The International Fellowship for Outstanding Researchers )}
        \thanks{Yasin Almalioglu is with the Computer Science Department, University
        of Oxford, Oxford, UK {\tt yasin.almalioglu@cs.ox.ac.uk}}
        \thanks{Mehmet Turan, Kutsev Bengisu Ozyoruk, Kagan Incetan and Guliz Irem Gokceler are with the Institute of Biomedical Engineering, Bogazici University, Turkey {\tt\{mehmet.turan,bengisu.ozyoruk,
        kagan.incetan,guliz.gokceler\}@boun.edu.tr}} 
        \thanks{Abdulkadir Gokce and Muhammed Ali Simsek are with the Electrical and Electronics Engineering, Bogazici University, Turkey 
        {\tt \{abdulkadir.gokce,ali.simsek\}@boun.edu.tr}}
        \thanks{Nicholas J. Durr is with the Department of Biomedical Engineering, Johns Hopkins University (JHU), Baltimore, MD {\tt ndurr@jhu.edu}}  
        \thanks{Kivanc Ararat is with the Department of Computational Engineering, Friedrich Alexander Universit{\"a}t Erlangen-N\"{u}rnberg, Germany {\tt\footnotesize  kivanc.ararat@fau.de} }
        \thanks{Faisal Mahmood is with the Department of Pathology, Harvard Medical School, Boston, MA {\tt\footnotesize faisalmahmood@bwh.harvard.edu} }
}


%

\markboth{Journal of \LaTeX\ Class Files,~Vol.~14, No.~8, August~2015}%
{Shell \MakeLowercase{\textit{et al.}}: Bare Demo of IEEEtran.cls for IEEE Journals}

\maketitle

\thispagestyle{plain}
\pagestyle{plain}

\begin{abstract}
Although wireless capsule endoscopy is the preferred modality for diagnosis and assessment of small bowel diseases, the poor camera resolution is a substantial limitation for both subjective and automated diagnostics. Enhanced-resolution endoscopy has shown to improve adenoma detection rate for conventional endoscopy and is likely to do the same for capsule endoscopy. In this work, we propose and quantitatively validate a novel framework to learn a mapping from low-to-high resolution endoscopic images. We combine conditional adversarial networks with a spatial attention block to improve the resolution by up to factors of $8\times$, $10\times$, $12\times$, respectively. Quantitative and qualitative studies performed demonstrate the superiority of EndoL2H over state-of-the-art deep super-resolution methods DBPN, RCAN and SRGAN. MOS tests performed by 30 gastroenterologists qualitatively assess and confirm the clinical relevance of the approach. EndoL2H is generally applicable to any endoscopic capsule system and has the potential to improve diagnosis and better harness computational approaches for polyp detection and characterization. Our code and trained models are available at {\color{blue} \textit{\href{https://github.com/CapsuleEndoscope/EndoL2H}{https://github.com/CapsuleEndoscope/EndoL2H}}}.

\end{abstract}


%
\IEEEpeerreviewmaketitle

\section{Introduction}
\IEEEPARstart{M}{inimally} invasive capsule endoscopy has become the preferred diagnostic modality for small bowel diseases since last decade. Despite successful clinical adoption, capsule endoscopy is still regarded to have a limited diagnostic yield due to interpretation errors which can be attributed to a variety of different factors including passive and uncontrolled motion, post-procedural assessment, sparse sampling of the organ due to mechanical or power limitations, and low resolution images due to capsule camera limitations \cite{park2019recent}. Several randomized controlled trials have shown the benefit of high resolution endoscopy for invasive endoscopic procedures \cite{pellise2008impact}. A recent assessment of five independent studies found that the incremental yield of high definition colonoscopy for the detection of any polyp was 3.8\% \cite{subramanian2011high}. Recent success in deep learning has enabled the automated and objective analysis of endoscopy including depth estimation \cite{mahmood2018unsupervised,mahmood2018deep}, polyp detection and characterization. On the other hand, recent work has also shown that low resolution can have a significant effect on the diagnostic performance of such algorithms \cite{ryai112}. Thus, there is a clear need for methods that can enhance resolution of capsule endoscopes for both subjective and objective analysis. 

\begin{figure}
	\centering
	\includegraphics[width=\columnwidth]{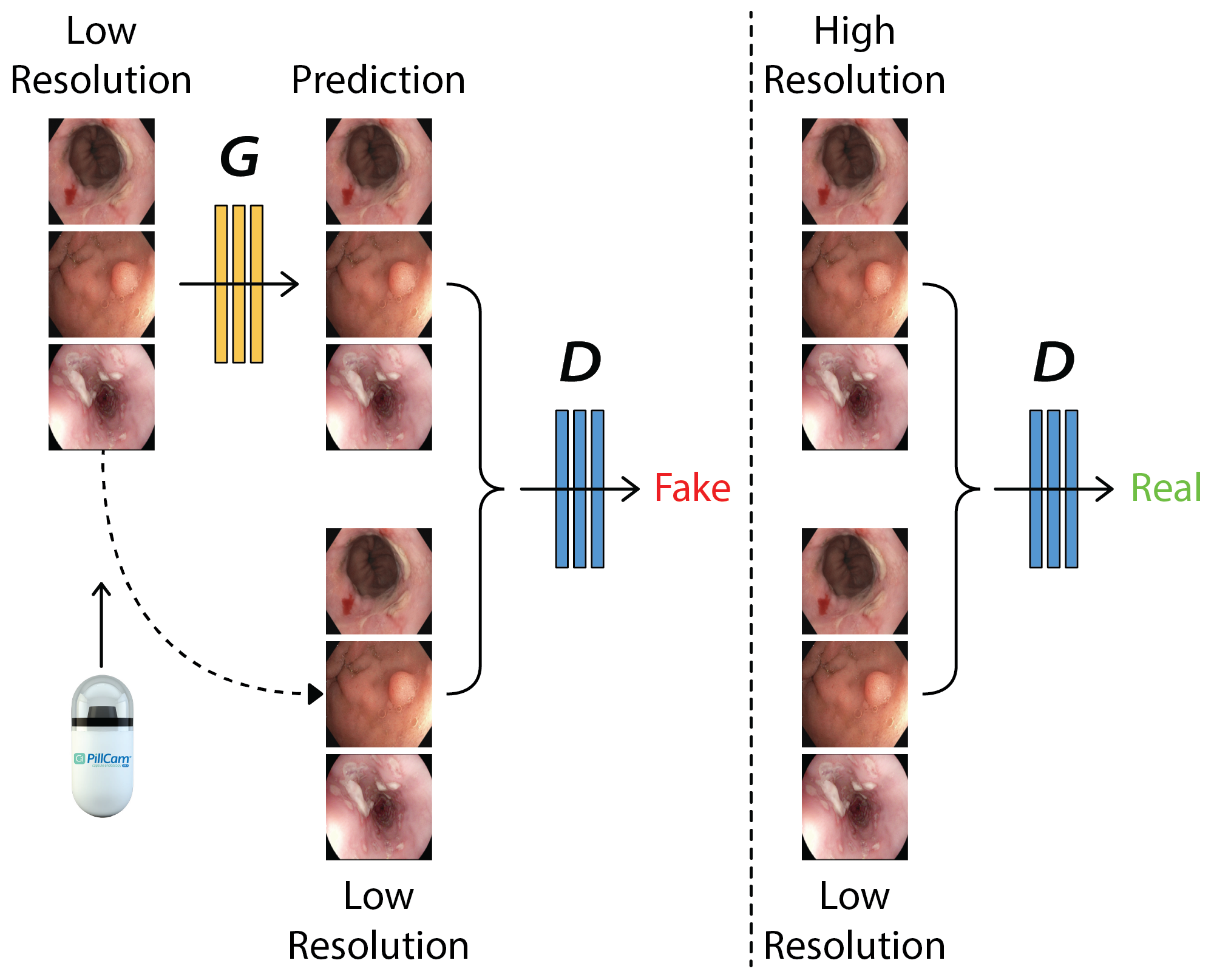}
	\caption{\textbf{System Overview.} A conditional GAN with an embedded spatial attention unit to map low resolution(LR) endoscopic images to diagnostically relevant high resolution(HR) endoscopic images. Unlike an unconditional GAN, both the generator and discriminator observe the input LR images.}
	\vspace{-4mm}
	\label{fig:summary}
\end{figure}

{\color{blue}Enhancing the resolution of images by increasing the size of the optics and the sensor array is not always a feasible solution since reasonable cost and critical space considerations are prohibitive for many endoscopic applications.} To address this issue, computer vision community has been developing a collection of algorithms known as super-resolution, which are used for generating high-resolution images from lower-resolution imaging systems. Enhanced image quality will likely lead to better disease/abnormality detection, region segmentation, 3D reconstruction, visual odometry, etc. \cite{turan2018deep, fischer2004capsule, nakamura2008capsule}. The ability to overcome fundamental resolution limits using super-resolution technique, which is a highly challenging process of reconstructing high resolution (HR) counterparts from low resolution (LR) camera outputs, has recently shown significant progress, potential and capability in numerous areas of medical imaging. Moreover, deep learning-based end-to-end super-resolution has demonstrated considerable success for a variety of imaging modalities \cite{ben2010super, kouame2009super, akhtar2010single, robinson2017new} in last decade. This paper proposes a deep super-resolution approach for capsule endoscopy images. The main idea is summarized and depicted in Fig. \ref{fig:summary}. 
{\color{blue}Our main contributions are as follows:

\begin{itemize}
\item  \textbf{Spatial Attention-based Super Resolution cGAN:} We propose a spatial attention based super-resolution cGAN architecture specifically designed and optimized for capsule endoscopy images. 

\item  \textbf{High fidelity loss function:} We introduce \textit{EndoL2H} loss which is a weighted hybrid loss function specifically optimized for endoscopic images. It collaboratively combines the strengths of perceptual, content, texture, and pixel-based loss descriptions and improves image quality in terms of pixel values, content, and texture. This combination leads to the maintenance of the image quality even under extremely high scaling factors up to, $10\times$-$12\times$.

\item \textbf{Qualitative and quantitative study:} We conduct a detailed quantitative analysis to assess the effectiveness of our approach and compare it to alternative state-of-the art approaches. We also conduct qualitative MOS tests performed by 30 gastroenterologists to evaluate the clinical applicability of the method. 

\end{itemize}
}

\section{Related Work}
Super-resolution algorithms can be classified based on various criteria such as the number of images used, transformation domain (spatial or frequency domain), color space, and so on \cite{Huang2015SR}. With recent advances in GPUs and dataset availability, learning-based super-resolution techniques have increasingly attracted attention \cite{Wang:15}. Glasner \textit{et al.} \cite{glasner2009super} make use of patch redundancies across scales within an image to achieve super-resolution. Huang \textit{et al.} \cite{huang2015single} extend self dictionaries by further allowing for small transformations and shape variations. Gu \textit{et al.} \cite{gu2015convolutional} proposed an approach which uses a convolutional sparse coding algorithm that works on the entire image instead of focusing only on overlapping patches resulting in perceptually better quality super-resolution images. Tai \textit{et al.} \cite{tai2006perceptually} combine an edge-directed SR algorithm relying on a gradient profile prior to the advantage of learning-based detail synthesis to regenerate higher-frequency image details. Gaussian Process Regression \cite{he2011single}, Trees \cite{salvador2015naive} or Random Forests \cite{schulter2015fast} may also aid the regression performance. In the context of super-resolution, deep learning methods try to learn the nonlinear mapping between low-resolution and high-resolution images in an end-to-end fashion utilizing neural networks such as Dong \textit{et al.} \cite{dong2016accelerating} making use of super-resolution Convolutional Neural Network (SRCNN) to cover high-frequency representation in a given low-resolution counterparts like sparse-coding-based strategies with the additional advantage of joint optimization. 

Superresolution techniques have been widely applied for medical images, as well. For MR images, as an example, Rueda \textit{et al.} in \cite{rueda2013single} uses HR and LR dictionaries that are learned from MRI. However, the fact that network performs well in the training  and not in the testings, indicates an overfit. Mahapatra \textit{et al.} in \cite{mahapatra2019image} uses progressive generative adversarial networks (P-GANs) so that more accurate detection of anatomical landmarks and pathology segmentation in medical images can be achieved. Numerous SR techniques have also been utilized to increase the quality of images acquired by low-resolution endoscopic cameras. In \cite{hafner2013pocs}, Hafner \textit{et al.}  propose an SR algorithm that is based on the projection onto convex sets (POCS) approach. They aim to reveal details such as mucosal structures that may not be seen on limited HD endoscope magnification. The construction is based on registration, fusion, and image restoration. In the work of K\" {o}hler \textit{et al.}  \cite{kohler2013tof} uses an SR technique based on a ToF (Time of Flight) sensor for range images. In the proposed work, an HR image is generated by multiple LR images with known subpixel displacements. This is done by estimating movements of the endoscope held by a surgeon using optical flow on RGB data from the endoscope camera. In \cite{rupp2007improving}, Rupp \textit{et al.} proposes an SR method in order to improve the calibration accuracy of flexible endoscopes. 

\section{Method}
\label{sec:method}
\begin{figure*} [!t] 
	\centering
	\includegraphics[width=\textwidth]{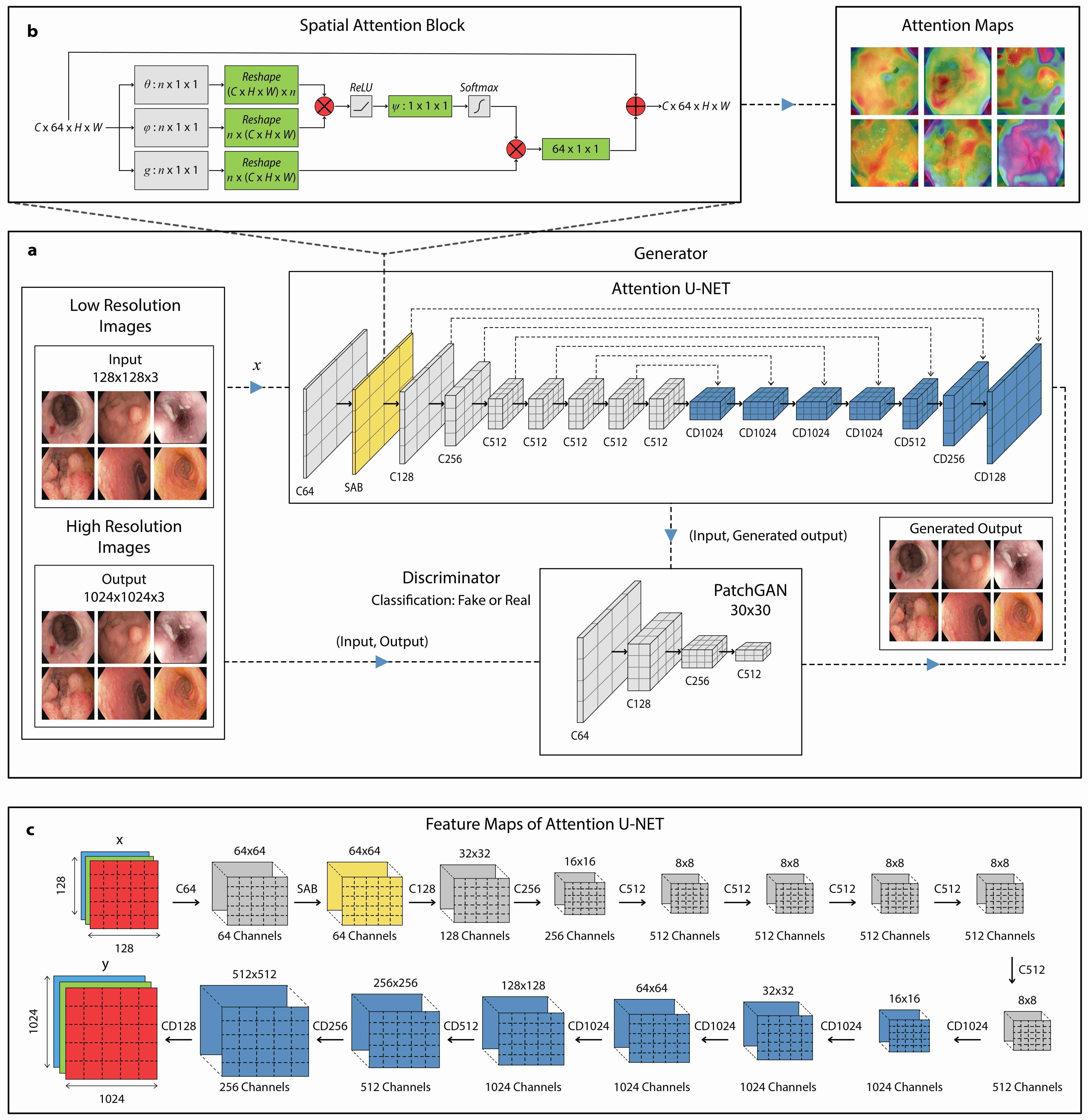}
	\caption{\textbf{Architecture Overview.}  \textbf{a} Overall system architecture of EndoL2H super-resolution framework. A low resolution input image is fed to the generator that creates an estimated high resolution counterpart, which is then served to the discriminator. The Markovian discriminator takes tuples of an LR input image and the corresponding HR image (real or generated), and tries to recognize whether the HR image is real or fake. Our generator is a derivation of the standard U-net with an additional SAB layer and it sequentially downsamples the input tensor by a factor of 2 (except the Spatial Attention Block) until the latent feature representation, and upsamples by a factor of 2 through the decoder layers. We use convolutional PatchGAN as a classifier which penalizes the structure in accordance with the image patch sizes (30$\times$30).  \textbf{b} The flow diagram of the spatial attention block (SAB) that is employed with a prospect of selectively focusing on clinically more relevant regions, its output images and gradient-weighted class activation mappings. \textbf{c} The feature maps of the Attention U-Net summarizing the applied filters and their input $\&$ output tensor sizes for 8$\times$ upscaling. The tensorflow of the low resolution image with dimension 128$\times$128$\times$3 throughout the Attention U-net ends up with 1024$\times$1024$\times$128. Similar to 8$\times$, tensor width and height evolves from 102$\times$102 to 1020$\times$1020 for 10$\times$ and from 85$\times$85 to 1020$\times$1020 for 12$\times$ throughout encoder-decoder layers.}
	\vspace{-7mm}
	\label{fig:architecture}
\end{figure*}
\subsection{Preliminaries for Super-Resolution Algorithms}
Super-resolution is the name of the technique applied to reconstruct  HR images from LR images. LR images $I^{LR}$, are modeled as the output of the following degradation process:
\begin{equation}
    I^{LR} = \mathcal{D} (I^{HR} ; \delta),
\end{equation}
where $I^{HR}$ denotes an HR image, $\mathcal{D}$ represents a degradation mapping function, and $\delta$ stands for the parameter set of the degradation process. The degradation process (i.e., $\mathcal{D}$ and $\delta$) is unknown and LR images are provided as an input for the inverse degradation function \cite{Wang2019}. The goal is to recover the corresponding HR image ${I}^{HR}$ from the LR image $I^{LR}$, so that ${I}^{SR}$ converges to the ground truth HR image $I^{HR}$, following the process:
\begin{equation}
    {I}^{SR} = \mathcal{F} (I^{LR} ; \theta_F),
\end{equation}
where $\mathcal{F}$ is the super-resolution model and $\theta_F$ represents the parameters of $\mathcal{F}$. The unknown degradation process can have been affected by various factors such as defocusing, compression artefacts, anisotropic degradations, sensor noise, speckle noise, etc, making superresolution task highly complex. Accordingly, in this study, we model the degradation as a combination of several distortion effects:
\begin{equation}
    \label{eq_degradation_impl_combine}
    \mathcal{D} (I^{HR} ; \delta) = (I^{HR} \otimes \kappa) \downarrow_r + n_\varsigma, \{\kappa,r,\varsigma\} \subset \delta,
\end{equation}
where $I^{HR} \otimes \kappa$ represents the convolution between a blur kernel $\kappa$ and the ground truth HR image $I^{HR}$, $\downarrow_r$ is a downsampling operation with the scaling factor $r$ (e.g., bicubic
interpolation with antialiasing), and $n_\varsigma$ is an additive white Gaussian noise with standard deviation $\varsigma$. The objective of the super-resolution process is as follows:
\begin{equation}
    \hat{\theta_F} = \mathop{\arg \min}_{\theta_F}( \mathcal{L}^{SR} (I^{SR}, I^{HR}) + \lambda \Phi_F (\theta_F)) ,
\end{equation}
where $\mathcal{L} ({I}^{SR}, I^{HR})$ denotes the loss function between the ground truth image $I^{HR}$ and the generated SR image $I^{SR}$, $\Phi_F (\theta_F)$ is the regularization term and $\lambda$ is the variance/bias trade-off parameter. The proposed architecture, EndoL2H, illustrated in Fig. \ref{fig:architecture}, consists of
four parts: a generator, a spatial attention module, a discriminator, and a hybrid objective function.
Let’s denote $I^{LR}$ and $I^{SR}$ as the low resolution input image and the superresolved output image, respectively. We define and train an attention U-Net generator network $G$ to reconstruct $I^{SR}$, $G_{\theta_G}$, parametrized by ${\theta_G}=\{W_{1:L}; b_{1:L}\}$ denoting the weights and biases of $L$-layer. In addition, we define a discriminator network $D$ to distinguish between ground truth HR and generated SR images. Ultimately, we define a hybrid loss function EndoL2H, $\mathcal{L}_{L2H}$, that iteratively optimizes $\theta_G$ with an objective of modelling distinct and desirable characteristics of the high definition endoscopic SR images.
Our objective function is as follow:
\begin{equation}
	\hat{{\theta}}_G = \arg\min_{\theta_G} \frac{1}{N} \sum_{n=1}^{N}{\mathcal{L}^{SR}(G_{\theta_G}(I^{LR}_n),I^{HR}_n)},
\end{equation}
where $I^{LR}_n$ is the input LR image with a corresponding $I^{HR}_n$ HR image, and $N$ is the size of the training dataset.

\subsection{Generator and Discriminator}
Our network design is inspired and modified from previously proposed pix2pix baseline approaches \cite{isola2017image, radford2015unsupervised}. To train the generator $G$ and the discriminator $D$ networks, we follow the well-known adversarial min-max optimization problem: \cite{isola2017image}: 
\begin{equation}
\label{eq:minmax}
\begin{split}
	\min_{\theta_G} \max_{\theta_D} ~& ( \mathbb{E}_{I^{HR}\sim p_\textrm{train}(I^{HR})} [ \log D_{\theta_D}(I^{HR}) ] + \\
	& \mathbb{E}_{I^{LR}\sim p_G(I^{LR})} [ \log (1-D_{\theta_D}(G_{\theta_G}(I^{LR})) ]).
\end{split}
\end{equation}
Equation \ref{eq:minmax} allows us to train a generative model $G$ with the goal of fooling the differentiable discriminator $D$ that is, in parallel, trained to distinguish superresolved endoscopic images from real HR endoscopic images. The training objective dictated to the generator, which is mainly based on a simple success criteria, that is just trying to fool the discriminator, enables the generator to learn reconstruction of perceptually superior SR images residing in the subspace, the manifold, of high resolution real endoscopic images and thus upon the training making the superresolved images difficult to be recognized by the discriminator. The architecture of the generator and discriminator networks are demonstrated in Fig. \ref{fig:architecture}-a. Our generator network is a derivation of the well-known U-net architecture with an additional spatial attention block. The input is processed through a series of convolutional layers that progressively downsample it until the latent feature representation is acquired and sequentially upsampled by the following decoder layers. Besides the bottleneck feature information, the low level information shared between input image and its high resolution counterpart needs to be passed across the layers, as well e.g., LR and HR images in an SR network share the location of prominent texture pixels which is for example very critical  in case of clinically abnormal and distinctive regions with pathological findings. To establish this information flow in an effective way, we adopt the general shape of the U-Net architecture \cite{ronneberger2015u} and add skip connections, concatenating all channels between each layer $i$ and the layer $n-i$, $n$ indicating the total number of layers. The main prospect of attention unit is to enable a selective focus on clinically abnormal, suspicious and more informative areas which inherently show more subtle and intense gradient changes based on textural and topographical alterations caused by the illness. Attention unit basically memorizes the spatial dependencies throughout the network and reveals regions that are more distinctive so that a higher importance can be attached therein. To accelerate convergence and reduce the risk of overfitting, we additionally apply batch normalization and decrease internal covariate shift \cite{ioffe2017batch}.
To be concrete, we perform the normalization for each mini-batch and train two extra transformation parameters for each channel to preserve the representation ability.
The batch normalization calibrates the intermediate feature distribution and mitigates the vanishing gradient problem, which allows us to use higher learning rates and be less careful about the network initialization.

{\color{blue}
Let \texttt{Ck} denote a stack of convolutional layers with $k$ filters, batch normalization and ReLU layer. \texttt{CDk} denotes a stack of convolutional layers with $k$ filters, batch normalization, dropout with a rate of $50\%$ and ReLU layer. All convolutions are $4 \times 4$ \cite{radford2015unsupervised, isola2017image} spatial filters applied with stride $2$. Convolutional layers in the encoder and the discriminator downsample the input tensor by a factor of $2$, whereas they upsample by a factor of $2$ in the decoder. The encoder-decoder architecture of the generator $G$ consists of:
\begin{itemize}
    \item {\bf Attention U-Net encoder:}\\
    \texttt{C64-SAB-C128-C256-C512-C512-C512-C512\\-C512}
    \item {\bf Attention U-Net decoder:}\\
    \texttt{CD1024-CD1024-CD1024-C1024-C512-\\C256-C128}
\end{itemize}
}
{\color{blue}We add a convolutional layer with $tanh$ non-linearity after the last layer in the decoder to map the input to the number of output channels, i.e. $3$. As an exception to the above notation, we omit batch normalization in the first \texttt{C64} layer of the encoder. All ReLUs in the encoder are leaky with a slope of $\alpha=0.2$, whereas they are non-leaky in the decoder  \cite{isola2017image}. The activation functions of the bottleneck layer are zeroed by the batch normalization, which effectively makes the innermost layer skipped. We fix this issue by removing the batch normalization from this layer. We use LeakyReLU activation with a slope parameter $\alpha=0.2$ and avoid max-pooling throughout the network, following the architectural guidelines summarized by Radford et al. \cite{radford2015unsupervised}.} $D$ network which contains four convolutional layers with an increasing number of filter kernels of size $3\times3$ is trained to solve the adversarial min-max optimization problem in Eqn. \ref{eq:minmax}. The number of features in each convolutional layer is increased by a factor of $2$ from $64$ to $512$ in parallel to the VGG network \cite{simonyan2014very}. We use strided convolutions to reduce the image resolution each time the number of features is doubled. {\color{blue} The resulting $512$ feature maps area followed by two fully-connected layers and a final sigmoid activation function to obtain a probability for super-resolution image classification.} 
In order to model high-frequency information in the input LR image, we analyze the structure in local image patches. 
Fig. \ref{fig:architecture}-c depicts the transformation of input image to SR output image across the generator layers in terms of pixel dimension change from encoder to decoder. Unlike standard conventional GANs, we design a PatchGAN-based discriminator architecture that penalizes structures at a patch level instead of the entire image. The $D$ network classifies if each $N \times N$ patch in an image is real or fake, where $N$ is the patch size that can be significantly lower than the input image size. We run $D$ convolutionally across the image and average all responses to provide the ultimate output. In addition, patch based design reduces the parameter size and makes the discriminator faster to evaluate. Moreover, patch-based discriminator effectively models the image as a Markov random field, which assumes the pixels to be separated by more than the patch size and to be conditionally independent \cite{li2016precomputed}. This Markovian assumption is explored in \cite{li2016precomputed}, which is also the common assumption in texture \cite{efros1999texture} and style \cite{efros2001image} modelling. {\color{blue}To optimize the patch size, patches of $15 \times 15$, $30 \times 30$,  $50 \times 50$ and  $70 \times 70$ pixels were evaluated and patches of $30 \times 30$ pixels outperformed other patch versions both in terms of PSNR and SSIM. Generally speaking, the observation was that larger patch sizes than $30 \times 30$ tend to generate overly-smooth tissue texture, while very small patch sizes are introducing artefacts.} The proposed discriminator architecture is as follows:
\begin{itemize}
    \item {\bf discriminator:}
    \texttt{C64-C128-C256-C512}
\end{itemize}
Following the guidelines of the original paper \cite{isola2017image}, we apply a convolution operation after the last layer to map to
a 1-dimensional output, followed by a sigmoid function.
As an exception, BatchNorm is not
applied to the first C64 layer. Finally, all ReLUs are leaky, with a slope of 0.2.
{\color{blue}

\subsection{Spatial Attention Block (SAB)}
Details of the integrated SAB module are shown in Fig.~\ref{fig:architecture}-b. The intuition behind adding an SAB module into the generator network is to selectively focus on clinically more relevant regions by assigning them a higher order of priority in terms of diagnostic decision taking. In our architecture, the SAB module is placed right after the first convolution layer. Inside the spatial attention block, there are three convolution layers decomposing the input data into three components: $\theta$, $\phi$ and $\mathit{g}$ (see Fig.~\ref{fig:architecture}-b). The attention mechanism is a non-local convolution process. For a given input $\mathbf{X} \in \mathcal{R}^{Cx64xHxW}$, this non-local operation is defined as follows:
\begin{equation}
\mathbf{Z}=\mathit{f}(\mathbf{X}, \mathbf{X}^T)\mathit{g}(\mathbf{X}),
\end{equation}
where \textit{f} represents the relationship of each pixel to another pixel on the input tensor \textbf{X}. The non-local operator extracts relative weights of all positions on the feature maps. For spatial attention block, we perform dot product operation on $\theta$ and $\phi$ for input $\mathbf{X}$, which is activated using ReLu function: 
\begin{equation}
    \psi(X) = \sigma_{relu}(\theta(\mathbf{X})\phi(\mathbf{X})),
\end{equation}
where $\sigma_{relu}$ is the ReLu activation function. The dot product of $\theta(\mathbf{X})\phi(\mathbf{X}^T)$ gives a measure of the input covariance, which is a degree of tendency between two feature maps at different channels.
We perform a matrix multiplication between $\psi$ and $g$ and activate the output matrix in $\mathit{softmax}$ function to extract the attention map $\mathbf{S}$. Lastly, an element-wise sum operation with the input $\mathbf{X}$ and attention map $\mathbf{S}$ gives the final output $\mathbf{E} \in \mathbb{R}^{Cx64xHxW}$:
\begin{align}
    \mathbf{S} &= \sigma_{softmax}(\psi(\mathbf{X})g(\mathbf{X})), \\
    \mathbf{E}_j &= \sum_i (s_{ij} x_{ij}), 
\end{align}
where $\sigma_{softmax}$ is the $\mathit{softmax}$ function.
There is an extra short connection between the input $\mathbf{X}$ to the output $\mathbf{E}$, making attention module to learn the residual mapping  more effectively. 
}
Unlike convolutional operations and channel attention approaches \cite{zhang2018image} that are mainly working patch-by-patch on the input image and extracting local features from the weighted input, the spatial attention model usually prefers to work on the entire image, making the underlying mechanisms more suitable for tasks like superresolution. As shown in Fig.~\ref{fig:architecture}-b, feature maps are vectorized so that \textit{i}-th vector represents the feature map at the \textit{i}-th channel. Their dot product extracts the auto-correlation of the input data and the softmax operation normalizes each vector to unit vectors, each corresponding to a principal axis of the input data. The dot product of \textit{g}(\textbf{X}) and the normalized vectors project the data to a new coordinate system. The output of the softmax is the global weight matrix that measures the importance of each feature map. Unlike PCA that uses the statistical correlation in a dataset to reduce the data dimensionality, the spatial attention module tries to capture and use the feature correlations across the entire spatial domain.

\subsection{Learning Objectives for Super-Resolution}
\label{sec:learning_strategies}

EndoL2H combines a derivation of standard GAN objective with a weighted sum of pixel, content and texture losses. Per-pixel loss measures the difference between ground truth and superresolution image on the pixel level, while the GAN discriminator mainly puts weight on modelling the high-frequency structures to ensure more realistic and useful output SR images. Unlike the standard GAN objective based on cross entropy, we define a GAN-loss based on least square error resulting in more stability during training and better convergence at the end of the training. In parallel to GAN-loss, content loss encourages perceptual similarity and consistency in low-frequency domains, while texture loss puts emphasis on recapturing texture information from the degradated input image. This hybrid loss function, EndoL2H loss, is specifically designed and empirically optimized for endoscopic type of images, making it unique for both standard and capsule endoscopic image superresolution.

\textbf{Pixel Loss.}
For numerical stability, we use a variant of the pixel L1 loss, namely Charbonnier loss \cite{bruhn2005lucas,lai2017deep}, given by:.
\begin{equation}
\label{eq:loss_cha}
    \mathcal{L}_{\text{Cha}} (I^{SR}, I^{HR}) = \frac{1}{hwc} \sum_{i,j,k} \sqrt{(I^{SR}_{i,j,k} - I_{i,j,k}^{HR})^2 + \epsilon^2}, 
\end{equation}
where $\epsilon$ (e.g., $1e-3$) is a small constant and h, w, c are height, width, color channel count of tensor. Pixel loss constrains the generated SR image $I^{SR}$ to be mathematically close enough to the ground truth HR image $I^{HR}$ in pixel level. Compared to L1 loss, L2 loss penalizes larger errors harder but is more tolerant to smaller errors, while L1 loss in general shows improved performance and convergence over L2 loss, making it more suitable for superresolution tasks.
The fact that PSNR (Section \ref{sec_iqa_psnr}) is highly correlated with pixel-wise difference, and that minimizing pixel loss directly maximizes PSNR, makes the pixel loss still widely used in super-resolution and related domains.
However, as mentioned previously, a well-known drawback of the pixel loss is its incapability to preserve the perceptual image quality \cite{johnson2016fei} and high-frequency information ending in overly smooth textures \cite{sajjadi2018frame}.

\textbf{Content Loss.}
In last years, content loss has been introduced into the super-resolution domain by numerous mainstream studies \cite{dosovitskiy2016generating,johnson2016perceptual}. Content loss basically evaluates semantic differences between SR and HR image using feature maps of pre-trained networks as semantic feature extractors. It essentially transfers high-level image descriptors into the SR domain to ensure content-consistency between LR and reconstructed SR image. This is quite crucial and of significant importance especially for medical image processing applications, since any non-existent and/or incorrect information introduced by the superresolution algorithm can mislead the clinician and is a potential for misdiagnosis. In terms of pre-trained, off-the-shelf feature extractors, VGG \cite{simonyan2014very} and ResNet \cite{he2016identity} layers are the ones that are most commonly preferred by deep learning community. In this work, we define the content loss based on the guidelines in \cite{gatys2015neural, bruna2015super, johnson2016perceptual}, using the ReLU activation layers of the pre-trained VGG network as described in \cite{simonyan2014very}. Let $\phi_{i,j}$ represent the feature map of the VGG network obtained by the $j$-th convolution layer before the $i$-th max-pooling layer. We define the VGG loss as the standard Euclidean distance between the feature representations $\phi_{i,j}$, the superresolved image $G_{\theta_G}(I^{LR})$ and the ground truth HR image $I^{HR}$:

\begin{equation}
\label{eq:loss_content}
	\begin{split}
	\mathcal{L}_{content} =
	\frac{1}{w_{i,j}h_{i,j}} & \sum_{x=1}^{w_{i,j}} \sum_{y=1}^{h_{i,j}} (\phi_{i,j}(I^{HR})_{x,y} \\
	& - \phi_{i,j}(G_{\theta_G}(I^{LR}))_{x,y})^2,	
	\end{split}
\end{equation}
where $w_{i,j}$ and $h_{i,j}$ are the dimensionality of the respective feature maps within the VGG network. The content loss of EndoL2H is a derivation of VGG-19 network that uses pre-trained weights on ImageNet as universal feature extractors, as described in Eqn. \ref{eq:loss_content}. 

\textbf{Texture Loss.}
Following the description introduced by\cite{gatys2015texture}, we model the texture of an image as the correlations between different feature maps of a layer and define it as the Gram matrix $G^{(l)} \in \mathcal{R}^{c_l \times c_l}$, where $G^{(l)}_{ij}$ is the inner product between the vectorized feature maps $i$ and $j$ on layer $l$:
\begin{equation}
    G^{(l)}_{ij}(I) = \operatorname{vec}(\phi_i^{(l)}(I)) \cdot \operatorname{vec}(\phi_j^{(l)}(I)) ,
\end{equation}
where  $\operatorname{vec}(\cdot)$ denotes a vectorization operation, and $\phi_i^{(l)}(I)$ represents the $i$-th channel of the feature maps on layer $l$. Based on these, the texture loss is described as:
\begin{equation}
\label{eq:loss_texture}
    \mathcal{L}_{\text{texture}} (I^{SR}, I^{HR} ; \phi, l) = \frac{1}{c_l^2} \sqrt{\sum_{i,j} (G^{(l)}_{i,j}(I^{SR}) - G^{(l)}_{i,j}(I^{HR}))^2}.
\end{equation}
By using the texture loss, we aim to create more realistic, visually more satisfactory and clinically more useful and trustable output SR images that are much closer to their HR counterparts \cite{sajjadi2018frame}.


\textbf{Adversarial Loss.}
\label{subsec:adv_loss}
{\color{blue}
Unlike the cross entropy-based GAN-loss proposed in SRGAN \cite{ledig2017photo}, we use adversarial loss based on least square error for more stability during training and better convergence at the end of the training, resulting in a more precise superresolution cite{ledig2017photo, mao2018effectiveness}, given by:
\begin{align}
\label{eq:loss_adv}
    \mathcal{L}_{\text{adv}} (I^{SR}, I^{HR} ; D) &= (D(I^{SR}))^2 + (D(I^{HR}) - 1)^2 .
\end{align}

For more details of the adversarial loss and training, the reader is referred to Section \ref{sec:method}.}

\textbf{EndoL2H Loss.} 
As mentioned earlier, EndoL2H loss is a hybrid loss function that is specifically designed and optimized for endoscopic type of superresolution applications. For that purpose, EndoL2H combines adversarial loss with L1-norm pixel loss, content loss and texture loss in an optimal way. The final formulation of EndoL2H loss is as follows:
\begin{equation}
\begin{split}
\label{eq:equation_4}
\mathcal{L}_{L2H}=&\alpha \mathcal{L}_{adv} +\\
&(1-\alpha)(1-\beta)(1-\gamma) \cdot \mathcal{L}_{Cha} +\\
&\gamma \cdot \mathcal{L}_{content} + \beta \cdot \mathcal{L}_{texture},
\end{split}
\end{equation}
where $\mathcal{L}_{Cha}$, $\mathcal{L}_{content}$, $\mathcal{L}_{texture}$ and $\mathcal{L}_{adv}$ are the loss components described in Eqn.~\ref{eq:loss_cha}, Eqn.~\ref{eq:loss_content}, Eqn.~\ref{eq:loss_texture} and Eqn.~\ref{eq:loss_adv}, respectively.
Variable $\alpha$ dynamically modulates the influence of the adversarial loss on EndoL2H, while $\beta$ and $\gamma$ are employed to modulate the influence of the texture and content loss, accordingly.

\section{Experimental setup and Evaluation metrics}
\label{sec:setup_eva}
The dataset used in this study is derived from the original Kvasir dataset \cite{pogorelov2017kvasir}. In terms of quantitative metrics, PSNR, SSIM, LPIPS and GMSD were applied. As per qualitative metric, a clinical MOS was performed, where 30 gastroenterogists voted on 15 randomly sampled images from the test dataset.
\begin{figure*}[!h]
	\centering
	\includegraphics[width=\textwidth]{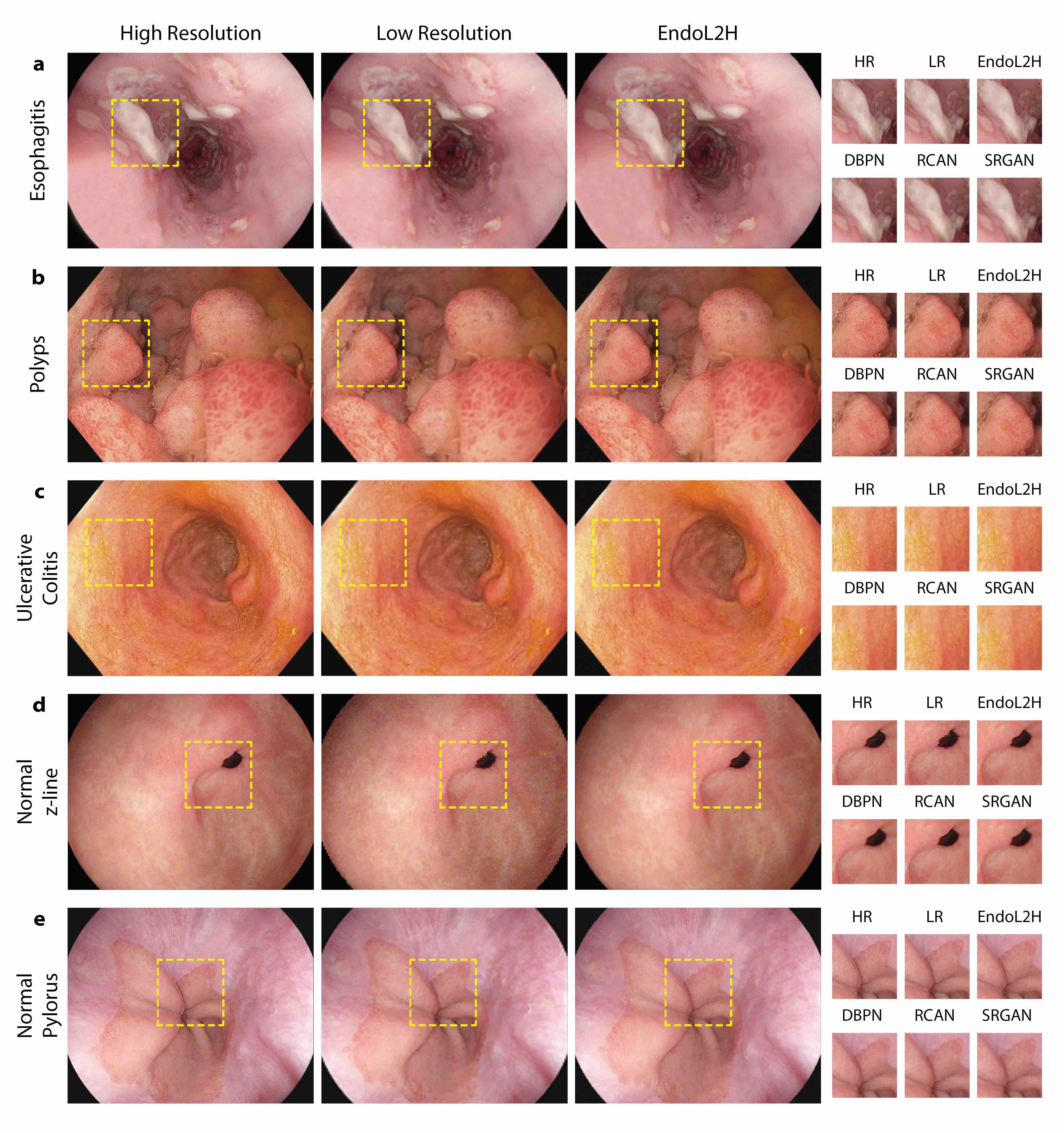}
	\caption{\textbf{Examples of superresolved endoscopic images} Super-resolution results for EndoL2H, DBPN, RCAN and SRGAN with cropped and zoomed regions inside the yellow squares on 8$\times$ upscale factor and for the five classes of the dataset. \textbf{a - c} are abnormal classes: esophagitis (inflammatory disease of esophagus), polyps (abnormal growth of mucous membrane of small and large intestine), and ulcerative colitis (inflammatory bowel disease), respectively. In general, compared to RCAN, SRGAN and Bicubic interpolation, EndoL2H and DBPN generate SR images with much sharper edges, finer texture details, less blur and artifacts introduced across the reconstructed image. Besides, EndoL2H successfully preserves and enhances clinically relevant and abnormal regions more faithfully to the ground truth HR images, which is of paramount importance and very critical in terms of effective, accurate and reliable disease diagnosis, afterwards.  In parallel to that, \textbf{d - e} are demonstrating the output SR images for healthy tissue images, such as for z-line (gastroesophageal junction images where the esophagus meets the stomach) and for pylorus images (a valve between stomach and duodenum).}
	\label{fig:crop_metrics}
\end{figure*}

\begin{figure*}
	\centering
	\includegraphics[width=\textwidth]{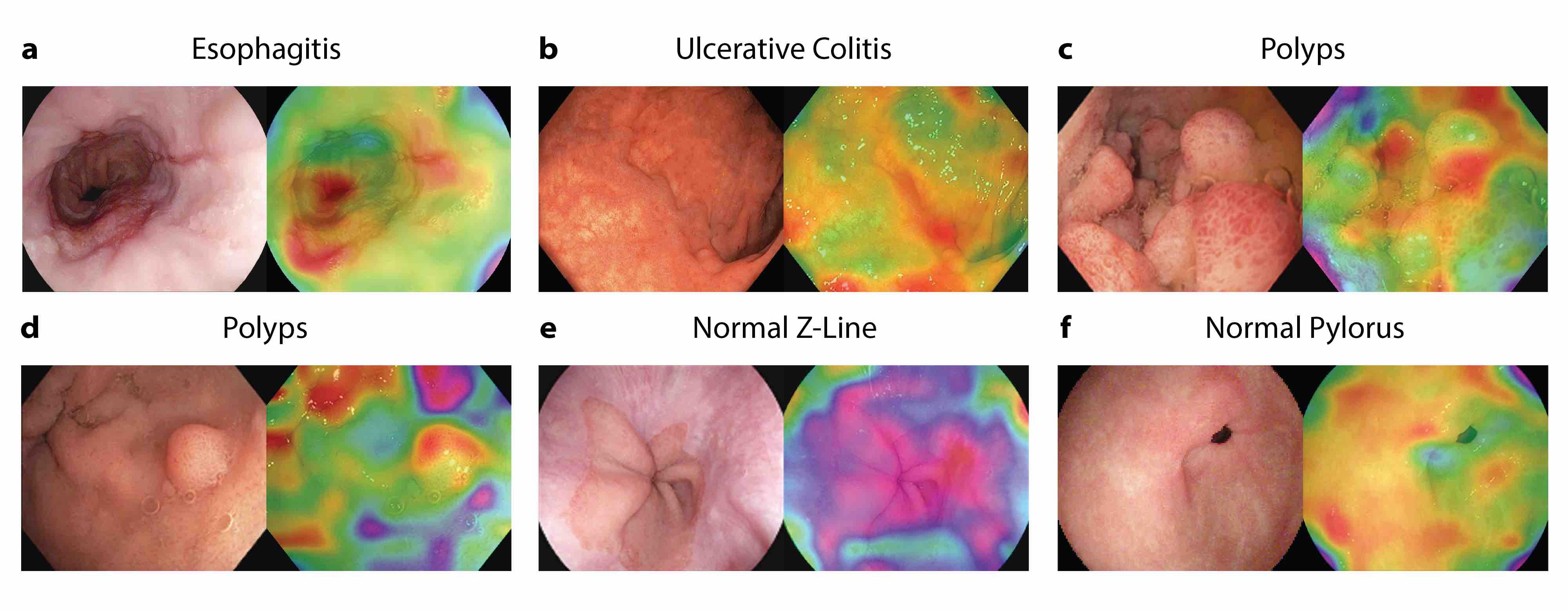}
	\caption{\textbf{Attention Maps.} Each image pair consists of LR images and corresponding SAB output visualised using Grad-CAM methodology \cite{gradCAM2017}. The main aim of SAB is to selectively put emphasis on more distinctive regions with unusual gradient alterations that inherently deserves attention as suspicious areas, in this case for example, regions with polyps, inflammatory tissue or normal tissue regions containing more texture and sharp edges: \textbf{a} esophagitis, \textbf{b} ulcerative colitis, \textbf{c-d} polyps, \textbf{e} normal Z-line and \textbf{f} normal pylors. As seen, SAB layer drives the attention of the superresolution network to areas with unusual structural and/or textural alterations, represented by color codes of higher wavelengths.}
	\vspace{-5mm}
	\label{fig:sr_samples}
\end{figure*}

\subsection{Dataset}
{\color{blue}
The original Kvasir dataset, consisting of $80,000$ endoscopic images, is recorded using various endoscopic equipments from Olympus (Olympus Europe, Germany) and Pentax (Pentax Medical Europe, Germany) at Vestre Viken Health Trust Hospital(VV) in Norway and acquired from different GI tract organs including bowel, stomach, esophagus, duodenum (v2) \cite{pogorelov2017kvasir}. Each of 10 folds consists of test and train sets containing eight different classes: dyed-lifted-polyps, dyed-resection-margins,
esophagitis, normal-cecum, normal-pylorus, normal-z-line, polyps and ulcerative-colitis. For this study, we removed image samples with large green annotations illustrating the position and configurations of the used endoscopic equipment. Remaining dataset consists of $21,220$ images with 1280x1024 resolution, containing images from classes esophagitis, normal-pylorus, normal-z-line, polyps and ulcerative-colitis. We created $I^{LR}$ by downsampling $I^{HR}$ ground truth image (RGB, $C=3$) using bicubic kernel with downsampling factors of $8\times$, $10\times$, and $12\times$ and an additional Gaussian filter as a blur kernel, respectively. This procedure augments data to faithfully represent existing state-of-the-art endoscopic cameras limited by low-resolutions due to energy and space limitations. For more details about the original Kvasir dataset  \cite{pogorelov2017kvasir} and the dataset we derived for EndoL2H study from the Kvasir dataset, the reader is referred to Appendix C.
 }
\subsection{Image Quality Assessment}
\label{sec_iqa}
In this section, we briefly introduce qualitative and quantitative evaluation metrics used in this work to analyse and critically compare EndoL2H, DBPN \cite{DBPN2018}, RCAN\cite{RCAN2018}, SRGAN \cite{SRGAN2016} and bicubic interpolation. As per quantitative metrics, PSNR \cite{PSNR2010}, SSIM\cite{SSIM2004}, LPIPS \cite{zhang2018perceptual} and GMSD \cite{xue2013gradient} were employed, while in terms of qualitative evaluation, a clinical MOS test was performed.
\subsubsection{Peak Signal-to-Noise Ratio}
\label{sec_iqa_psnr}
Peak signal-to-noise ratio (PSNR) is a full-reference image quality measure of pixel-wise similarity. It is defined by a maximum possible pixel value $L$ and the mean-squared-error $MSE$ between the SR and HR image. The formula is given as \cite{PSNR2010}:
\begin{align}
    \label{equ:mse}
    {\rm MSE}  &= \frac{1}{N} \sum_{i=1}^{N} (I^{SR}(i) - I^{HR}(i)) ^2 , \\
    {\rm PSNR} &= 10 \cdot \log_{10} (\frac {L^2} {\rm MSE}),
\end{align}
where $I$ is HR image and $\hat{I}$ is the SR image both consisting of $N$ pixels. Even though PSNR ignores structural and perceptual similarity, it is still a widely and commonly used image quality metric in superresolution domain and other image reconstruction applications. 

\subsubsection{The Structural Similarity Index}
\label{sec_iqa_ssim}
Structural similarity index (SSIM) is another full-reference image quality metric that takes into account structural composition of pixels, as well. By making use of luminance, contrast and structure values of SR and HR image; SSIM aims to measure perceived quality by human visual system in the range of values between 0 and 1 and is more sensitive to high frequency content -such as edge and textures in comparison to PSNR. For more details about SSIM, the audience is referred to the original paper \cite{SSIM2004}. 


\begin{figure*}[!h]
	\centering
	\includegraphics[width=\textwidth]{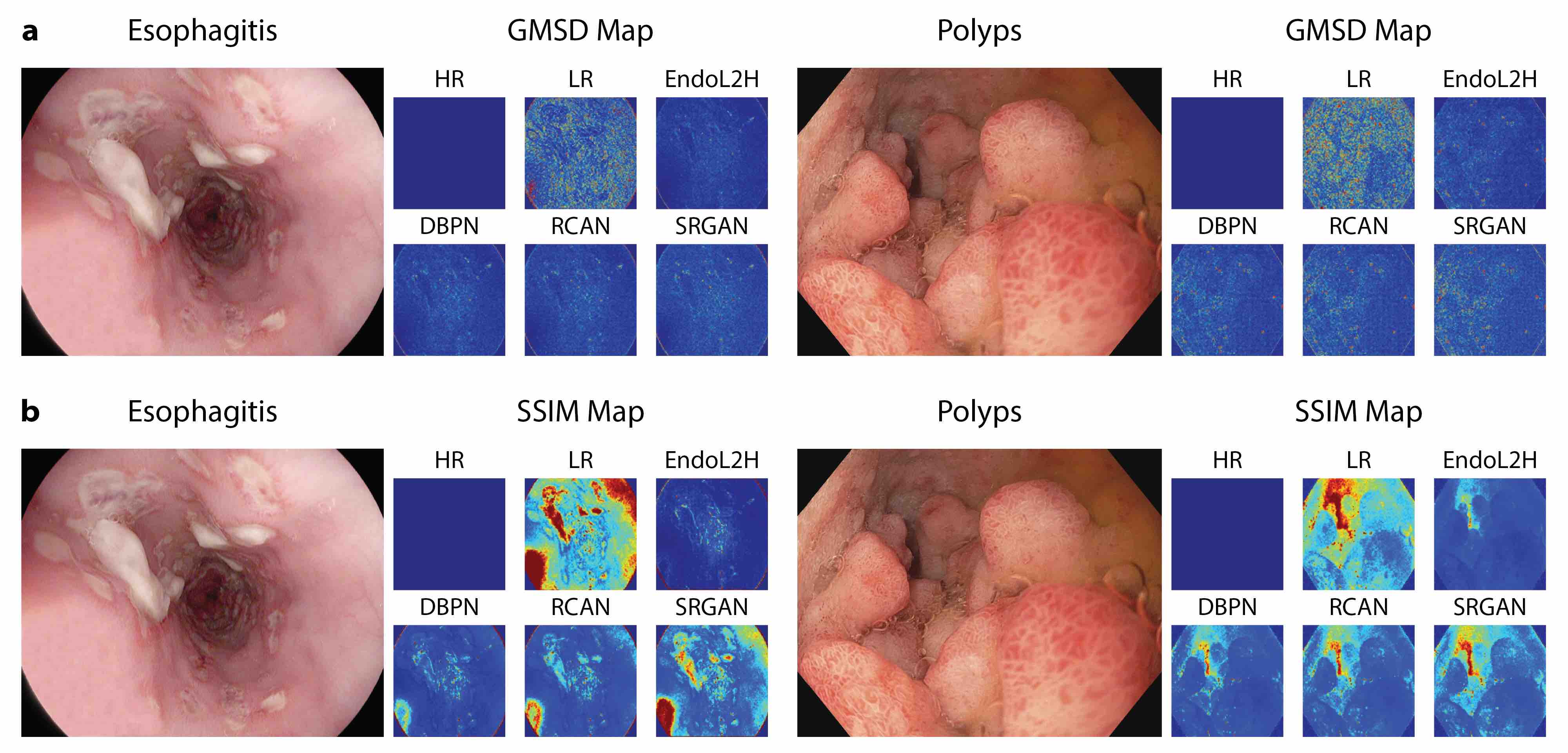}
	\caption{\textbf{GMSD and SSIM quality maps.} Results for the evaluations of image sets in Fig.  \ref{fig:crop_metrics} in terms of structural and gradient map similarity. \textbf{a } Resulting GMSD quality maps. High resolution image, low resolution image, results of SRGAN, DBPN, RCAN and EndoL2H are provided, respectively. Each point in the GMSD quality map represents the local GMSD  value  for 11$\times$11 Gaussian  window. Red  color denotes  higher GMSD  values indicating lower scores in terms of  gradient based similarity with the  original image  and blue color  represents  lower  GMSD  values indicating higher similarities with the original image in terms of gradient maps \textbf{b } SSIM heat maps. High resolution image, low resolution image, results of SRGAN, DBPN, RCAN and EndoL2H are provided, respectively.  Each  point  represents  the  local SSIM  value for 11$\times$11 Gaussian  window.  Red  color  denotes  lower SSIM  values, denoting a lower  structural  similarity  with the  original  image  and  blue color represent higher SSIM  values implying a higher structural similarity with the original image. As seen, EndoL2H outperforms DBPN, RCAN and SRGAN both in terms of SSIM- and GMSD- based similarity evaluations.
	}
	\label{fig:ssim_maps_new}
\end{figure*}

\subsubsection{Learned Perceptual Image Patch Similarity (LPIPS)}
\label{sec_iqa_ll}

While it is nearly effortless for humans to quickly assess
the perceptual similarity between two images, the underlying processes are thought to be quite complex. Despite that, most of the widely used perceptual metrics today, such as PSNR and SSIM, are rather simple and shallow functions failing to account for many nuances of human perception system. As an example, an image that has higher PSNR and SSIM scores can be more blurry than an image reconstructed by a GAN and has lower PSNR and SSIM scores \cite{ledig2017photo}. To circumvent this dilemma between mathematical and perceptual similarity interpretations, learned perceptual image patch similarity metrics have been introduced in recent years. (LPIPS) is one of these group of learned similarity metrics introduced by \cite{zhang2018perceptual}. In analogy to content loss, LPIPS basically utilizes feature maps of well-known and widespread baseline deep neural networks that were previously pre-trained on large and well-known datasets consisting of millions of image patches with corresponding human opinion scores in terms of perceptual distance. It than infers a perceptual similarity score between the high resolution and the superresolved image. In this work, version 0.1 of LPIPS repository with pre-trained AlexNet model was used with default input arguments. For more details about LPIPS, the reader is referred to the original paper \cite{zhang2018perceptual}.

{\color{blue}

\subsubsection{Gradient Magnitude Similarity Deviation (GMSD)}
\label{sec_iqa_gmsd}
Gradients on an image in general convey key and helpful information for many image processing applications, ranging from image analysis to image reconstruction and generation. Analysing the gradient information; underlying structural, textural and content related differences/similarities between images can be effectively recognized and revealed. Thus, as a final similarity metric, we used GMSD introduced by \cite{xue2013gradient} which is a member of gradient based similarity methods family. The image gradient magnitude is in general very responsive to small artifacts making it generally speaking a very effective similarity index for endoscopic type of images that inherently contain certain amount of local and small-sized structures, such as veins, specularities caused by organ fluids, and small critical clinical findings such as polyps and esophagitis. These differences cause different degradations in gradient magnitudes. Based on these considerations, the global variation of local quality degradation can be used as a reliable indicator of the image quality. GMSD basically compares pixel-wise gradient magnitude maps of the ground truth HR and superresolved SR image and sums them up to achieve a global gradient magnitude difference score between the compared SR and HR image. For further details of GMSD, the reader is referred to the original paper \cite{xue2013gradient}. 
}
\subsubsection{Mean opinion score (MOS) evaluations}
\input{tables/metrics_summary}
\label{sec_iqa_mos}
To further evaluate and compare SR performance of EndoL2H, DBPN, RCAN and SRGAN, we performed a clinical MOS evaluation in order to subjectively quantify the sharpness, suitability for diagnosis and detail retain level of the superresolved image by using votes of 30 gastroenterologists from various health institutions in Turkey.
{\color{blue} 15 images were randomly selected from test datasets containing three images from each of five examined classes: Esophagitis, normal-pylorus, normal-z-line, polyps and ulcerative-colitis. We asked gastroenterologists to assign an integer score to 15 output images from 1 (bad quality) to 5 (excellent quality) for each of three comparison metrics and SR images, respectively. To standardize the MOS procedure, same images were shown to each gastroenterologist.}

\section{Evaluations and Results}
\label{sec:eva_res}
This section contains details about optimization and inference procedure, gives comprehensive quantitative analyses for the proposed hybrid loss function and its variations, summarises performance evaluation results for the proposed and compared methods and finally examines and questions the effectiveness of the SAB module in the superresolution context. 

\subsection{Optimization and inference}
\input{tables/mos_results}
{\color{blue}
We trained and tested EndoL2H network on an NVIDIA® Tesla® V100 instance. The dataset was divided into three subsets: $17,220$ images for training, $2,000$ images for validation, and $2,000$ for testing. To avoid overfitting and to minimize sampling bias, the dataset was split into five folds with different train and test sets, keeping the validation set constant and starting the training each time from scratch to avoid cross-data leakage.
As per training protocol, EndoL2H was trained with $10^5$ update iterations at a learning rate of $10^{-4}$ and another $10^5$ iterations at a lower rate of $10^{-5}$. The optimization protocol loosely follows the guidelines described in Section~\ref{subsec:adv_loss} to train the generator $G$ and the discriminator $D$. Finally, we fine-tuned both together for another $2,000$ generator updates. In total, $200$ epochs were run using PyTorch and Adam optimizer \cite{kingma2014adam}, with momentum parameters $\beta_1=0.5$, $\beta_2=0.999$. 
}

\subsection{Ablation study for EndoL2H loss}
\label{subsec:loss}
{\color{blue}In this subsection, we investigate different loss combinations for EndoL2H network. Specifically, we analyse the following loss combinations:
\begin{itemize}
	\item EndoL2H loss, 
 	\item EndoL2H-w/o-C: EndoL2H loss without $\mathcal{L}_{content}$,
	\item EndoL2H-w/o-T: EndoL2H loss without $\mathcal{L}_{texture}$.
\end{itemize}
Quantitative results,  summarized in Table \ref{tab:metrics_summary}, indicate that EndoL2H-w/o-T produces perceptually less convincing results with smoother and coarser output SR images compared to EndoL2H-w/o-C, revealing the effectiveness of the texture loss for SR. We believe that this performance degradation is mainly caused by a competition between the content and the adversarial loss in the absence of texture loss. We further attribute small artifacts caused by superresolution procedure mainly observed in a minority of EndoL2H-w/o-C superresolved images, to those competing objectives. Moreover, in the case of $\mathcal{L}_{content}$, we additionally observe that using higher level VGG feature maps in general leads to higher quantitative similarity scores and more satisfying SR images compared to lower level VGG feature maps.

An important challenge when using hybrid loss functions is the difficulty of control parameter optimization that is ultimately determining which loss component should contribute at what degree to the learning process. To optimally determine the EndoL2H loss weights, we randomly sub-sampled 3,000 images from the dataset and generated 10 different hyper-parameter sets for $\alpha$, $\beta$ and $\gamma$. As seen in Table \ref{tab:loss_params} in Appendix F, parameter setting $\alpha=0.35$, $\beta=0.20$ and $\gamma=0.15$ shows the best performance in terms of PSNR, SSIM, LPIPS and GMSD metrics. In the rest of the evaluations, this parameter setting was used. 
The difference between total hyper-parameter count in between compared four methods also result in run time variations. Run-time comparison for four methods for all up-scaling factors can be seen in Table \ref{tab:run_time} of Appendix G.   }

\begin{figure*}
	\centering
	\includegraphics[width=\textwidth]{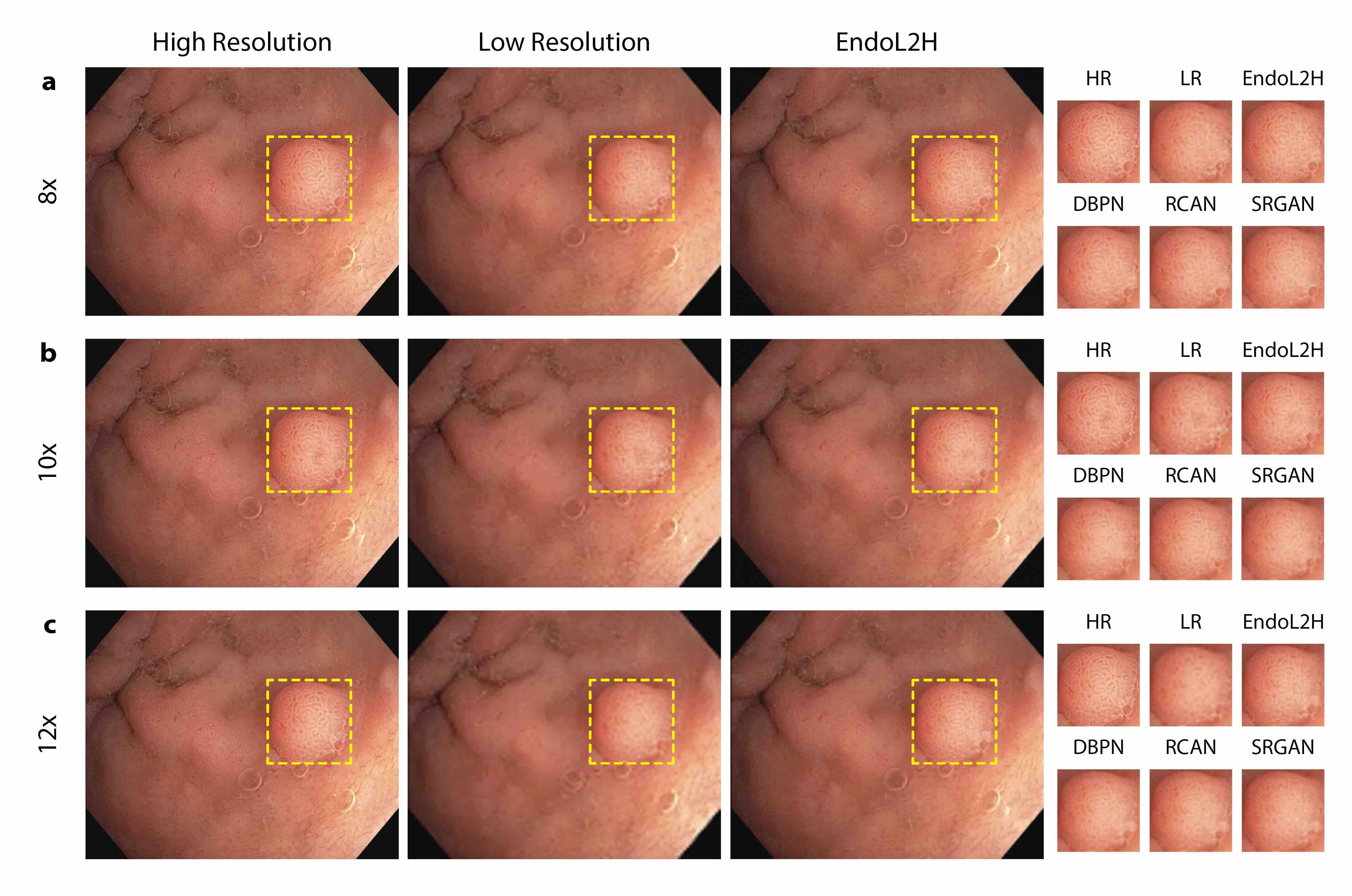}
	\caption{\textbf{Superresolution performance of the algorithms in case of upscale factors 10$\times$ and  12$\times$ compared to 8$\times$.}  Low resolution, high resolution and superresolved images for EndoL2H, DBPN, RCAN and SRGAN can be seen in the figure. Although DBPN and RCAN are also able to alleviate blurring and over-smoothing artifacts to some degree, they tend to produce unpleasing artifacts for 10$\times$ and  12$\times$. For example, in case of cropped parts that contain intensive vein textures, EndoL2H is able to preserve contours around the polyp shape and vein details on the polyp tissue more faithfully in comparison to DBPN and RCAN. At some small patchs, DBPN and RCAN suffer from moderate blurring artifacts and they fail to recover vein details. However, EndoL2H loss combination keeps fighting and successfully preserves the superresolved image quality up to a certain degree even under highly challenge and extreme scaling factor such as $12\times$.}
	\label{fig:scale_comparison}
\end{figure*}

{\color{blue}
\subsection{Performance evaluations for EndoL2H network}
\label{subsec:msebased}
We analyze the performance of algorithms on image sets of five different classes; namely, esophagitis, normal-pylorus, normal-z-line, polyps and ulcerative-colitis. To truthfully reveal the superresolution performance, we use four common quantitative assessment scores, which are PSNR, SSIM, LPIPS and GMSD. Fig.\ref{fig:crop_metrics} a-e demonstrate ground truth HR and the corresponding output SR images for each method, implying that EndoL2H is able to superresolve perceptually more pleasing and clinically more informative SR images in comparison to DBPN, RCAN and SRGAN (The reader is referred to see Fig. \ref{fig:results2} in Appendix B for further examples). Generally speaking, EndoL2H and DBPN are much better at reproducing sharp edges and small details compared to RCAN and SRGAN. Mean and standard deviation of these metrics achieved by EndoL2H, DBPN, RCAN, SRGAN and bicubic interpolation are given in Table \ref{tab:metrics_summary}, respectively. In most of the cases, EndoL2H results in the highest mean scores for PSNR and SSIM and lower mean scores for LPIPS and GMSD distances. The reader is referred to the Table \ref{tab:metrics_psnr}, \ref{tab:metrics_ssim}, \ref{tab:metrics_lpips} and \ref{tab:metrics_gmsd} in Appendix D to see class-dependent SR scores for the evaluated methods).

\begin{figure}
	\centering
	\includegraphics[width=\columnwidth]{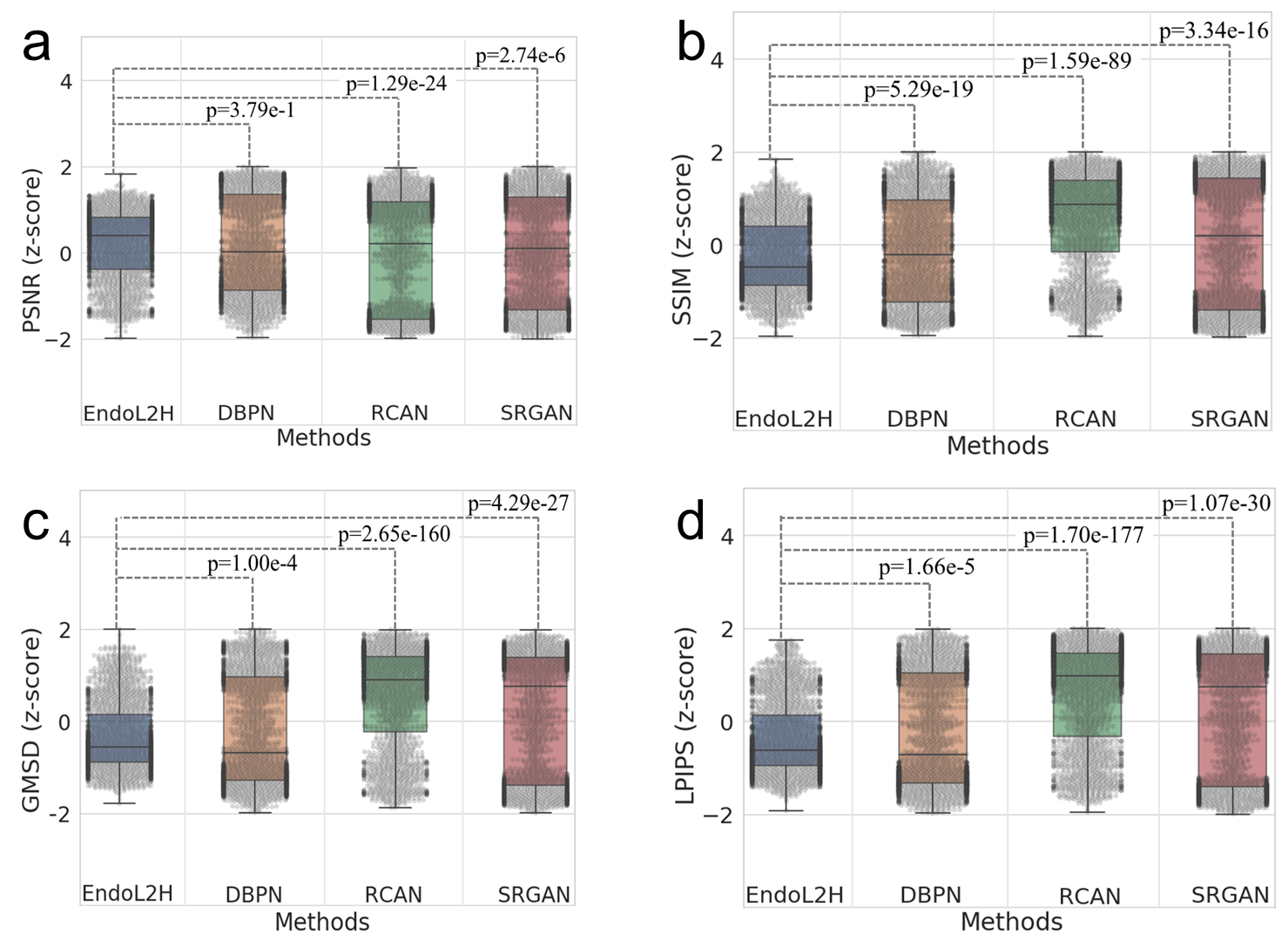}
	\caption{\textbf{Z-score Distributions} For EndoL2H, DBPN, RCAN and SRGAN; the z-score distributions and interquartile ranges with respect to \textbf{a} PSNR, \textbf{b} SSIM, \textbf{c} LPIPS and  \textbf{d} GMSD are given in 8$\times$ up-scaling. For all metrics, z-scores of SRGAN accumulated around two means due to its performance variety on different classes of Kvasir dataset. Generally speaking, the most Gaussian-like distribution is created by EndoL2H which is an indicator of homogeneous performances across five classes. Besides, the p-values acquired via the Wilcoxon Signed Rank Test can be seen at the top of each box plot. For further information, the reader is referred to the Table \ref{tab:wilcoxon_test}.}
	\label{fig:pvalues}
\end{figure}

Fig. \ref{fig:ssim_maps_new} shows SSIM and GMSD heatmaps for the compared methods and the ground-truth images, respectively. Both heatmaps imply that EndoL2H achieves the most structural and gradient magnitude similarities with respect to the ground truth counterparts. More comparison cases are given in Appendix B Fig. \ref{fig:gmsd_maps_app} and Fig. \ref{fig:ssim_maps_new_app}. 

As a further investigation, we trained and tested compared methods for extreme up-scaling factors. Table \ref{tab:metrics_summary} shows quantitative comparisons  for larger up-scaling factors ($10\times$ and $12\times$). Although EndoL2H and DBPN perform slightly better than RCAN and SRGAN, the output SR images in general indicate lack of some underlying structures and tissue details, see Fig. {\ref{fig:scale_comparison}. Decrease in PSNR $\&$ SSIM and increase in LPIPS $\&$ GMSD values in Table \ref{tab:metrics_summary} support this observation, as well. Results in Table \ref{tab:metrics_summary} also reveal importance of $\mathcal{L}_{content}$ and $\mathcal{L}_{texture}$ for higher upscaling factors as the performance of EndoL2H decreases in the absence of both losses. Better performance of EndoL2H-w/o-C compared to EndoL2H-w/o-T in most of the quality metrics implies that texture loss contributes more than the content loss to SR in accordance to the evidences we obtained for 8$\times$ .  

Since for some of the similarity index evaluations, DPBN results are very close to EndoL2H results, we additionally perform statistical significance analysis and show the distribution of z-scores for EndoL2H, DBPN, RCAN and SRGAN in terms of PSNR, SSIM, LPIPS and GMSD metrics (see Fig. \ref{fig:pvalues}). For that, we employ non-parametric Wilcoxon Signed Rank Test to check if there is an evidence of statistical significance that the EndoL2H superresolves LR images better than DBPN, RCAN and SRGAN in terms of PSNR, SSIM, LPIPS and GMSD \cite{wilcoxon2011}. The null hypothesis assumption in this case is claiming that there is no difference between the distribution of EndoL2H results and each of the remaining methods DBPN, RCAN and SRGAN for any of the given similarity metrics. The resultant difference between EndoL2H and any of the compared algorithm on an image pair $(I_{m}^{SR},I_{m}^{HR})$ is:
\begin{equation}
    \Delta_{m}^{M} = M_{EndoL2H}(I_{m}^{SR},I_{m}^{HR})-M_{algo}(I_{m}^{SR},I_{m}^{HR})
    \label{eqn:MetricDiff}
\end{equation}
where algo $\in$ \{DBPN, RCAN, SRGAN\}, M $\in$ \{PSNR, SSIM, GMSD, LPIPS\} and the image index m $\in$ \{1,...,n\} for n is the number of images in the test dataset including all of the five fold sub-test datasets. The absolute  differences are sorted from
$min(|\Delta| _{1}^M, ... , |\Delta|_{n}^M)$ to $max(|\Delta|_{1}^M, ... , |\Delta|_{n}^M)$. The rank order is assigned to image pairs starting from 1 to n where the pair with minimum absolute difference is matched with 1 and the maximum is n. Finally, replacing the sign of rank order with the sign of $\Delta_{m}^{M}$, we obtain $W_{m}$. In summary, to compute the differences caused by the methodology on each image pair, the sign of differences are attached to the rank and summed to obtain test statistics, W, formulated between EndoL2H and any of the compared methods as:

\begin{equation}
    W = \sum_{m=1}^{n} W_{m}.
    \label{eqn:W}
\end{equation}
The distribution of all possible sum of signed ranks approximates the normal distribution with mean $\mu_{W}=0$ and standard deviation:
\begin{equation}
    \centering
    \sigma_{W}=\sqrt{\frac{n(\ n+1 )\ (\ 2n+1 )\ }{6}}.
\end{equation}
Thus, z-score of the underlying distrubution is given by:
\begin{equation}
    \centering
    z_{W}={\frac{W - \mu_{W}}{\sigma_{W}}}={\frac{W}{\sqrt{[n(\ n+1 )\ (\ 2n+1 )\ ] /6}}}.
\end{equation}

For the detailed p-value formulation for this distribution, see \cite{originalWilcoxon1945}.
Based on the p value results in Table \ref{tab:wilcoxon_test}, we can reject the null hypothesis for DBPN, RCAN and SRGAN with strong evidence for each metric.  

\begin{table}[!h]
\color{blue}
    \centering
    \caption{\color{blue} P-values of the statistical significance analysis performed to compare 8$\times$ up-scaling performances of EndoL2H, DBPN, RCAN and SRGAN in terms of PSNR, SSIM, GMSD and LPIPS. All of the P-valaus  are $\leq$ 0.05 providing the evidence against the null hypothesis and statistically proving the superiority of EndoL2H over the remaining three methods.}
    \label{tab:wilcoxon_test}
\begin{tabular}{|l|l|l|l|}
 \hline
\textbf{Metric}	& \textbf{DBPN}	& \textbf{RCAN}	& \textbf{SRGAN}	\\ \hline
PSNR $\uparrow$			& $3.79e-1$		& $1.29e-24$		& $2.74e-6$		\\ \hline
SSIM $\uparrow$			& $5.29e-19$		& $1.59e-89$		& $3.34e-16$		\\ \hline
GMSD $\downarrow$			& $1.00e-4$		& $2.65e-160$		& $4.29e-27$		\\ \hline
LPIPS $\downarrow$		& $1.66e-5$		& $1.70e-177$		& $1.07e-30$		\\ \hline
\end{tabular}
\end{table}

MOS results, listed in Table \ref{tab:mos_results} further support our quantitative and qualitative findings. While EndoL2H obtains highest scores in terms of suitability for diagnosis and detail level of super-resolved images with 4.69 and 4.75 mean scores, DBPN outperforms EndoL2H and remaining methods in terms of sharpness. Class dependent and detailed results of MOS evaluation can be seen in Appendix D Table \ref{tab:detailed_mos}.

\subsection{Superresolution performance in the case of clinically relevant abnormalities}
The effect of superresolution on small artifacts and clinically relevant abnormalities are of paramount significance since preserving clinically abnormal regions and improving the perceptual quality are of critical importance in terms of clinical relevance of the SR operation. Consequently, we particularly analyze the skills of the compared methods in terms of retaining and elevating clinically relevant small findings during super-resolution process without any distortions. Fig. \ref{fig:crop_metrics} shows examples of clinical findings and SR outputs by the evaluated methods, including regions with pathological findings, such as esophagitis, polyp and ulcerative colitis. Although DBPN and RCAN are able to alleviate blurring and over-smoothing artifacts to some degree, they also tend to produce unpleasing artifacts. As an example, for the polyp image in Fig. \ref{fig:scale_comparison} of Appendix B, where the cropped part is full of veins and textures, EndoL2H preserves polyp contours and fine details with curved lines more faithfully to the original HR images in comparison to DBPN and RCAN. At small patches, DBPN and RCAN suffer from moderate level of blur artifacts and they fail to recover vein details (see Fig. \ref{fig:results_scale_comparison} of Appendix B for further examples).

\subsection{Ablation studies for Spatial Attention Block}
In order to make the network focus on clinically more relevant regions, a Spatial Attention Block (SAB) module was integrated  between the first and the second convolution layer of the generator network. Using such an attention mechanism, we aim to exploit the inter-dependencies among feature channels and contextual information outside the local image regions with an expectation to preserve both the low-frequency and high-frequency information from the input LR space in a better way. 
Fig. \ref{fig:sr_samples} a-f exemplifies polyps, ulcerative-colitis, esophagitis, normal-z-line and normal pylorus images from Kvasir dataset, and their combination with the output attention maps obtained by the SAB module of EndoL2H and visualized via GradCAM methodology. As seen, attention module makes EndoL2H to focus on clinically more suspicious regions, which inherently differ from their surrounding tissue in terms of texture and topology, ending in a prioritization of these regions in terms of superresolution quality compared to smooth and low-textured areas. 
Quantitative ablation analyses for SAB and its role in superresolution analysed in Table \ref{tab:ablation_sab} further support our qualitative observations and reveal the effectiveness and usefulness of spatial attention mechanism for super-resolution. Even in the case of $10\times$ and $12\times$, EndoL2H achieves acceptable similarity scores, indicating its persistent effort to preserve the superresolution quality in extreme conditions. Finally, Fig. \ref{fig:it_metrics} of Appendix B compares the training performance of EndoL2H with and without the SAB layer. Training curves of EndoL2H with the SAB imply a moderate increase in the performance and result in better convergence rates at the end, manifesting the effectiveness of attention unit mechanism for training SR networks.

\input{tables/ablation_sab}
}
}

\section{Conclusion}
\label{sec:discussion}

In this paper, we introduced EndoL2H, a deep SR approach, specifically designed and optimized for endoscopic capsule images. The proposed approach combines content, texture and pixel loss with adversarial loss to train a conditional GAN. Comprehensive quantitative analyses and clinical MOS evaluations prove the success of EndoL2H and show its superiority over DBPN, RCAN and SRGAN. As a future direction, we plan to analyse effects of EndoL2H in various endoscopy related vision tasks such as disease classification, region segmentation, depth and pose estimation from endoscopic images.

\section*{Acknowledgment}
\label{sec:acknowledgement}
We would like to express deep gratitude to Abdülhamid Obeid for his valuable reviews. Mehmet Turan, Kutsev~Bengisu~Ozyoruk, 
Kagan Incetan and Guliz Irem Gokceler are especially grateful to Technological Research Council of Turkey (TUBITAK) for International Fellowship for Outstanding Researchers.   

\if CLASSOPTIONcaptionsoff
  \newpage
\fi
\small\bibliographystyle{IEEEtran} 
\small\bibliography{mybibfile}

\newpage

\onecolumn

\section*{Appendix A. Green Channel Comparisons}

\begin{figure*}[!h]
	\centering
	\includegraphics[width=\textwidth]{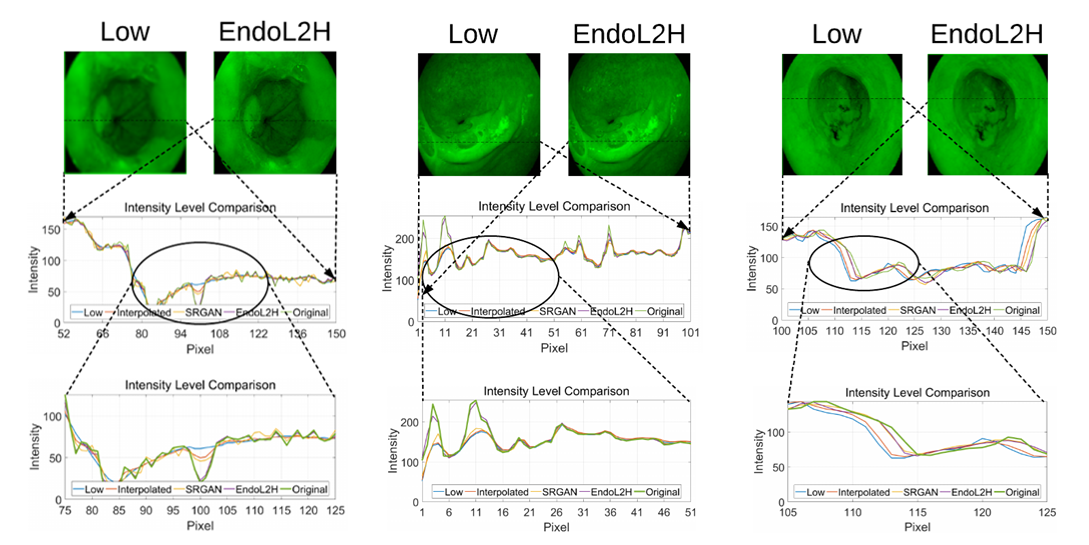}
	\caption{Green channel of different sample of images are shown for a pair of RL and SR generated by EndoL2H. Intensity profile comparisons of green channel for various tuples are also provided.}
	\label{fig:greencombinedchannels}
\end{figure*}

\section*{Appendix B. Image Quality Assessment}
\begin{figure*}[!h]
\includegraphics[width=\textwidth]{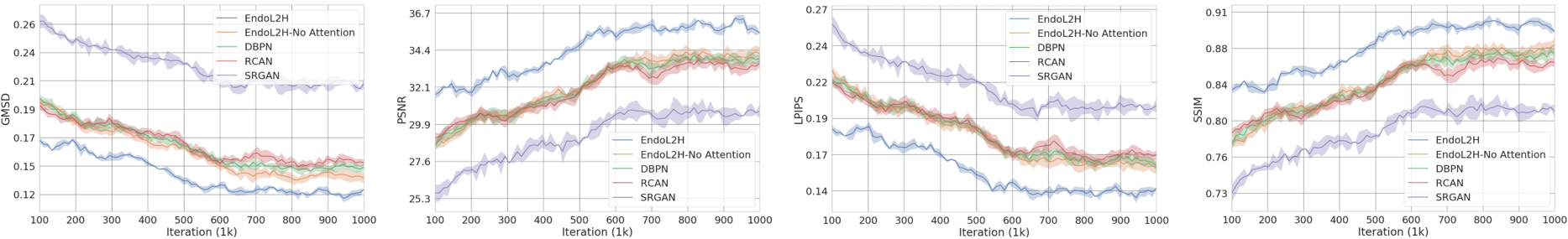} 
\caption{Training performance of EndoL2H w/wo SAB and other methods in terms of GMSD, PSNR, LPIPS and SSIM.}
\label{fig:it_metrics}
\end{figure*}

\begin{figure*}[!h]
	\centering
	\includegraphics[width=\textwidth]{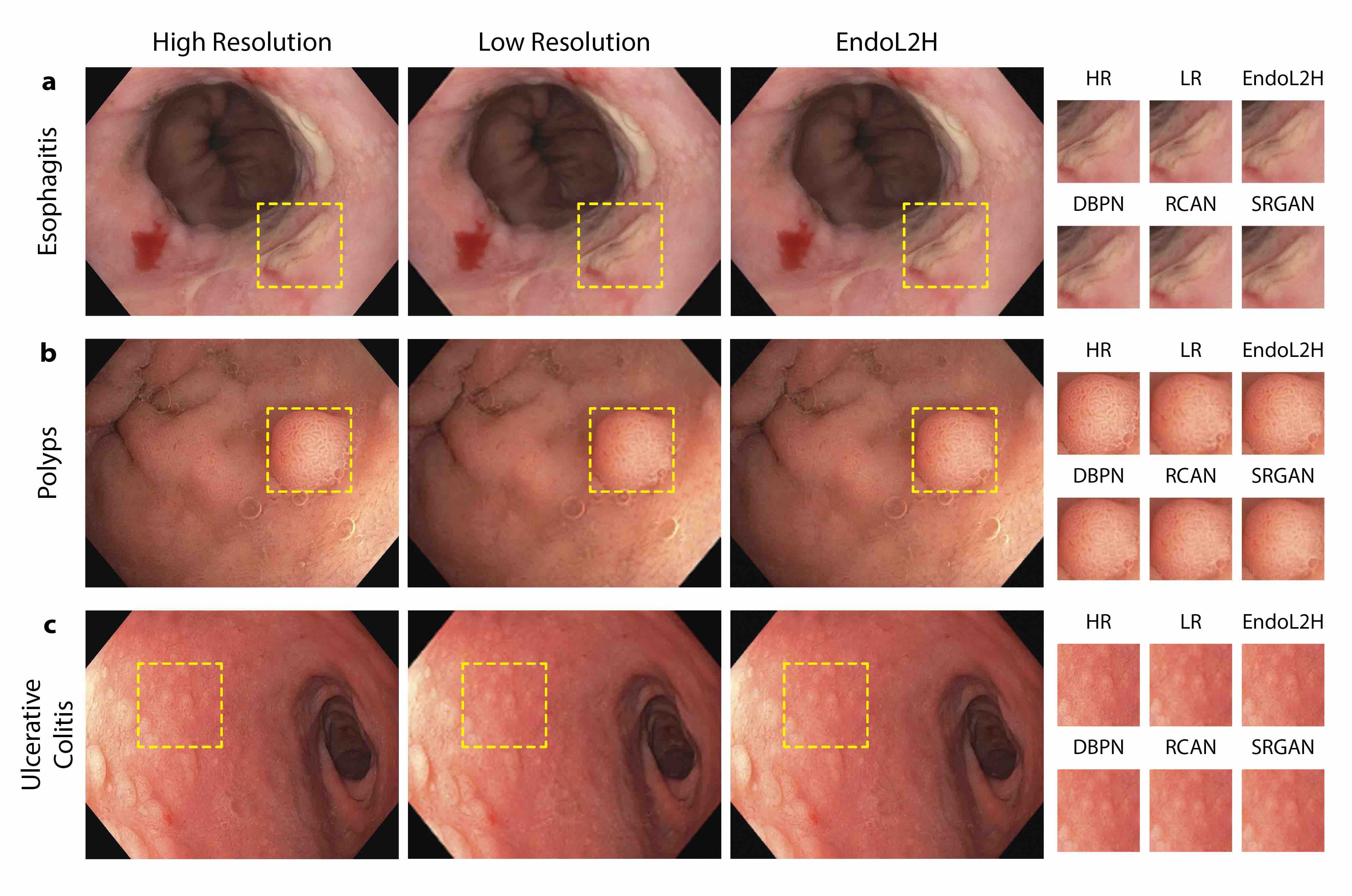}
	\caption{\textbf{Result Comparisons.} Performance Comparison. Super-resolution results of EndoL2H, DBPN, RCAN and SRGAN with the zoomed versions of the areas depicted inside the yellow squares on $8\times$ enlargement with input (LR) and groud truth (HR) images are represented for 3 abnormal classes of the Kvasir dataset: esophagitis which is basically inflammatory disease of esophagus, polyps abnormal growth of mucous membrane of small and large intestine, and ulcerative colitis similarly inflammatory boweldisease, respectively. In general, EndoL2H and DBPN generate much sharper edges than other approaches with less artifacts across  an  image.  EndoL2H  successfully  preserves  and  enhances  clinically  relevant  findings  at  the  regions  of  abnormalities which is crucial for an accurate detection after SR..}
	\label{fig:results2}
\end{figure*}
\begin{figure*}[!h]
	\centering
	\includegraphics[width=\textwidth]{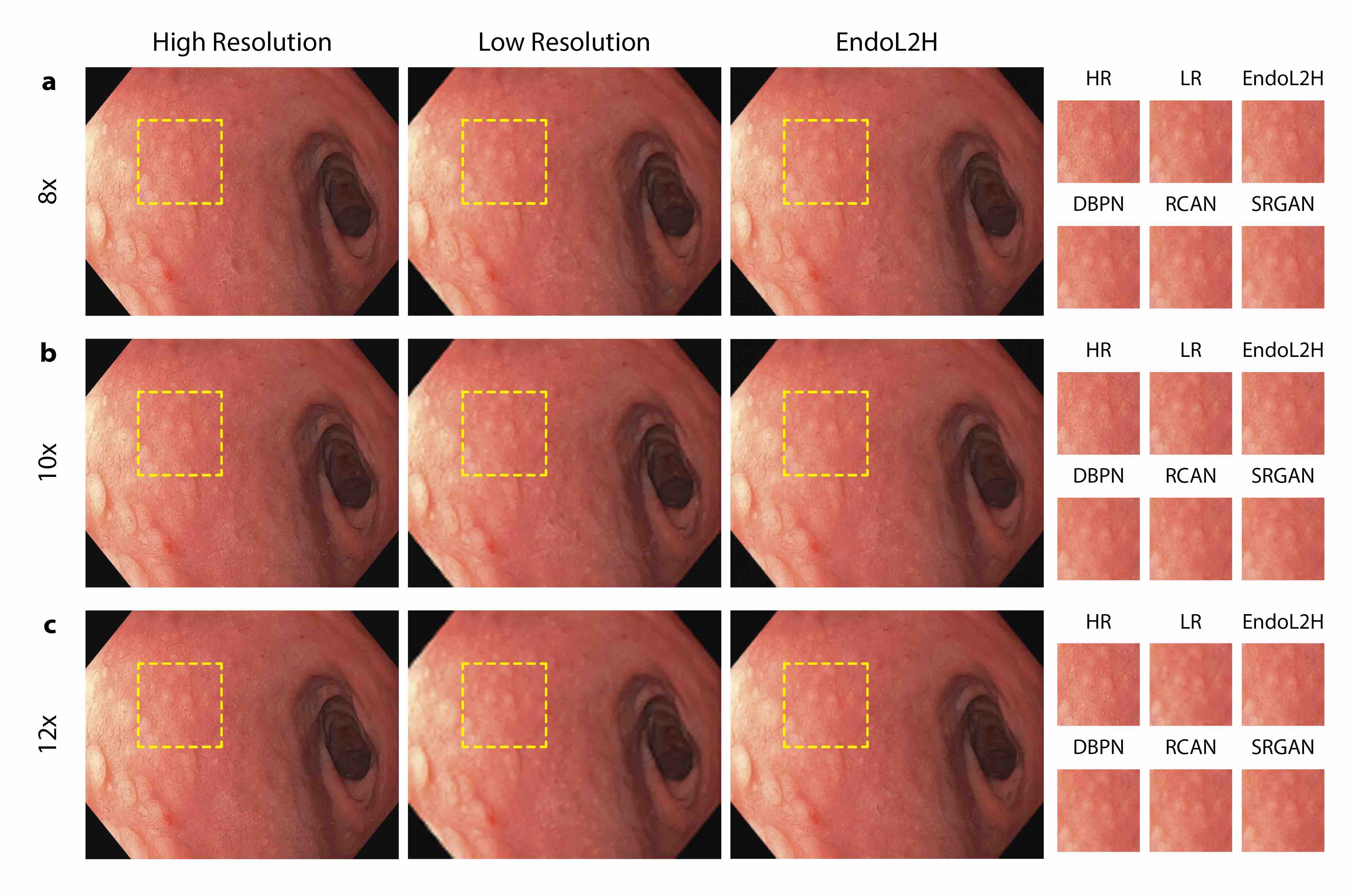}
	\caption{\textbf{Scale Comparison.} The low $\&$ high resolution and super resolution images of EndoL2H, DBPN, RCAN and SRGAN are given for \textbf{a} 8$\times$, \textbf{b} 10$\times$ and  \textbf{c} 12$\times$ for ulcerative colitis. Although DBPN and RCAN are able to alleviate blurring and over-smoothing artifacts to some degree, they also tend to produce unpleasing artifacts for all scaling factors. The cropped part of the polyp image is full of vein textures, EndoL2H preserves polyp contours and lines more faithfully in comparison to DBPN and RCAN. At some small patchs, DBPN and RCAN suffer from moderate blurring artifacts and they fail to recover vein details. However, EndoL2H loss combination leads to the maintenance of the image quality even under high scaling factors up to, $10\times$-$12\times$. }
	\label{fig:results_scale_comparison}
\end{figure*}

\begin{figure*}[!h]
	\centering
	\includegraphics[width=\textwidth]{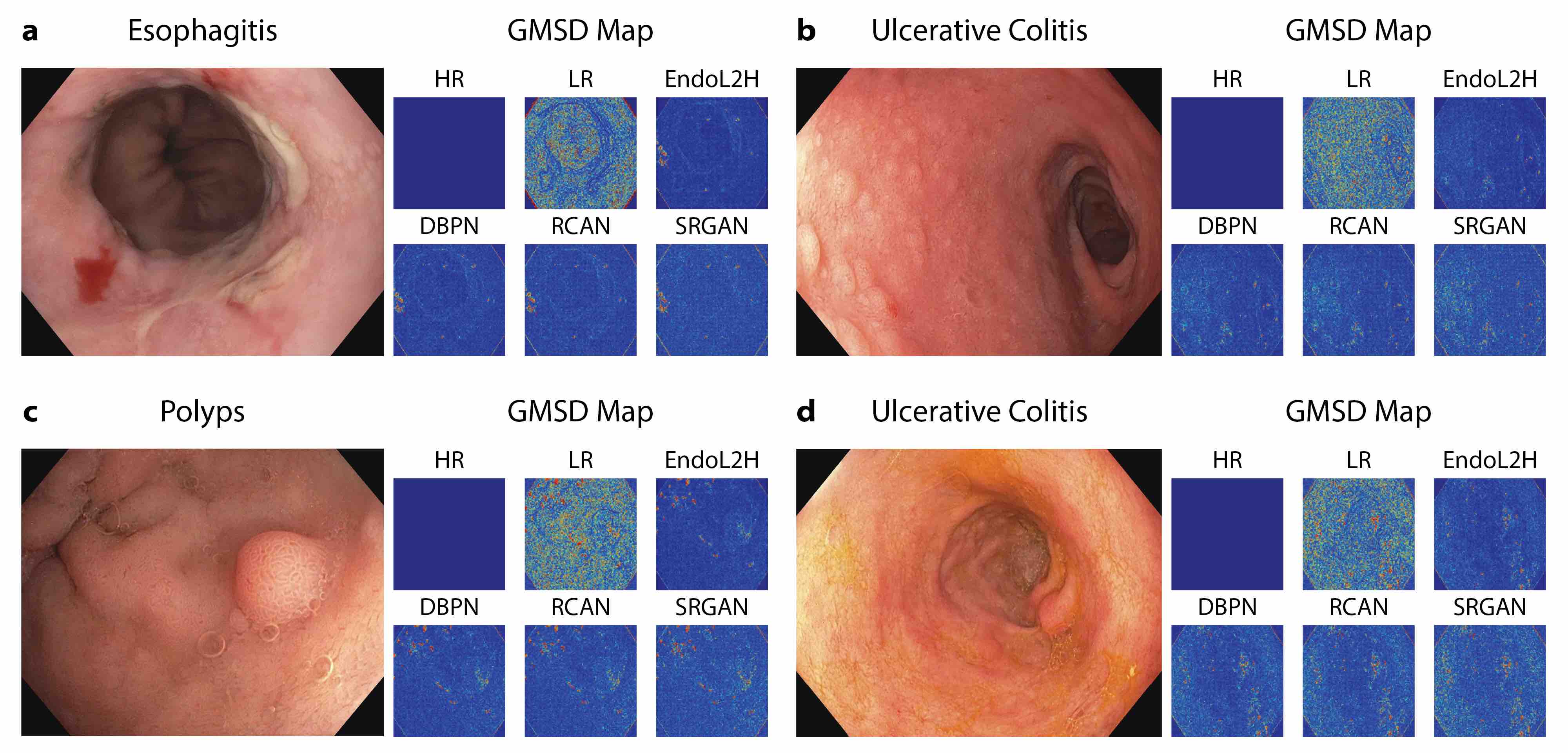}
	\caption{\textbf{GMSD Maps.} Resulting GMSD maps for 4 cases are shown. High resolution image, its corresponding low resolution version, results of SRGAN, DBPN, RCAN and EndoL2H are shown respectively. Each  point  represents  the  local  GMSD  value  for 11$\times$11 Gaussian  window.  Red  color  denotes  higher GMSD  values  (low  structural  similarity  with  the  original  image)  and  blue  color  represent  low  GMSD  values  (high  structural similarity with the original image)}
	\label{fig:gmsd_maps_app}
\end{figure*}
\begin{figure*}[!h]
	\centering
	\includegraphics[width=\textwidth]{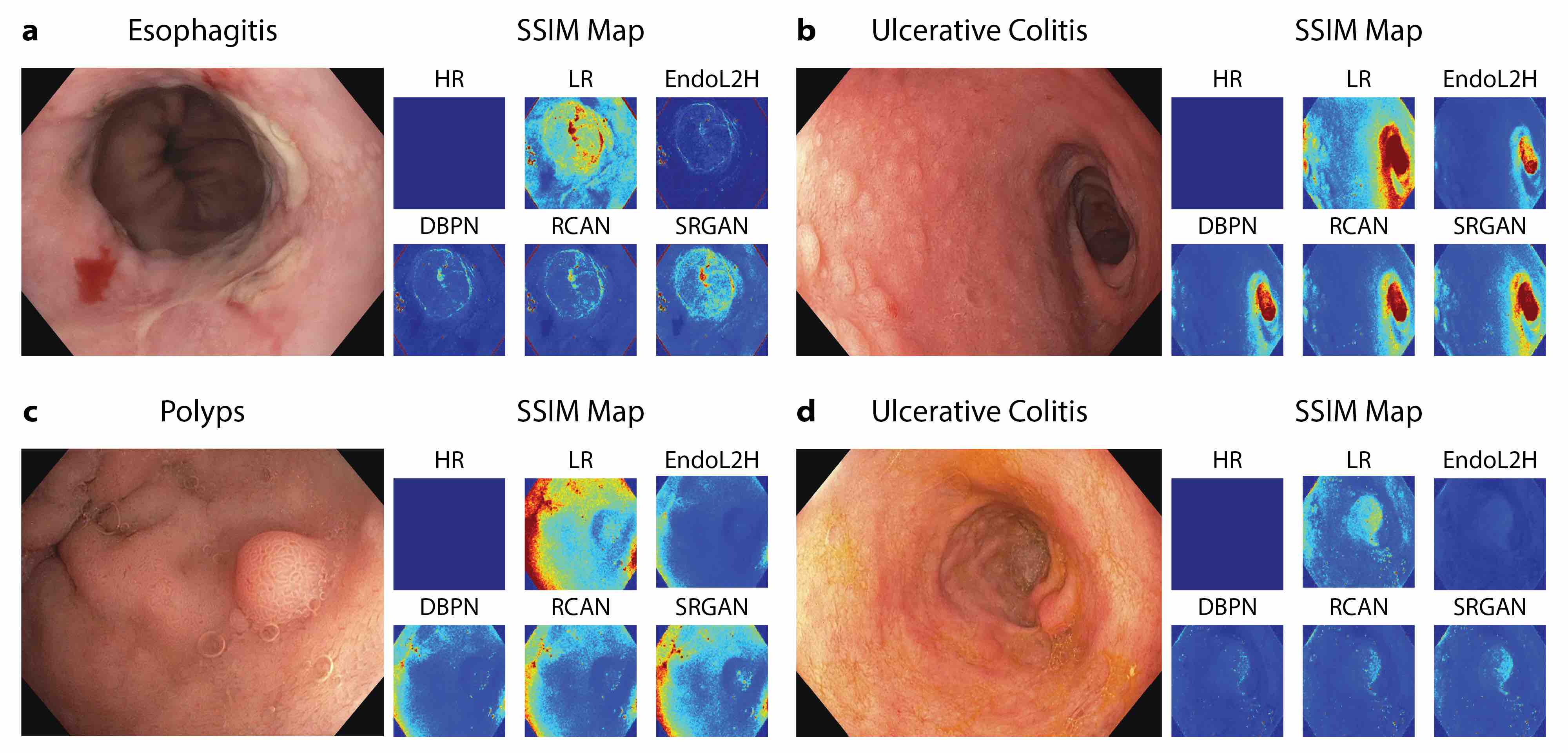}
	\caption{\textbf{SSIM Maps.} SSIM heat maps for 4 cases are shown. High resolution image, its corresponding low resolution version, results of SRGAN, DBPN, RCAN and EndoL2H are shown respectively.  Each  point  represents  the  local  SSIM  value  for 11$\times$11 Gaussian  window.  Red  color  denotes  lower SSIM  values  (low  structural  similarity  with  the  original  image)  and  blue  color  represent  high  SSIM  values  (high  structural similarity with the original image).}
	\label{fig:ssim_maps_new_app}
\end{figure*}

\clearpage
\newpage
\section*{Appendix C. Original Dataset}
\label{app:dataset}

The original Kvasir dataset folds v2 (size 25.3 GB) archive contains $80,000$ images from inside the gastrointestinal (GI) tract for automatic detection of diseases to improve medical practice and refine health care systems. 
The dataset contains train and test splits, which are ready to perform 10-folds cross-validation of the multi-class classification algorithms. 
Images are stored in separate folders, which are named according to the fold numbers and class names. 
Image files are encoded using JPEG compression. 
The encoding settings can vary across the dataset, which reflect \textit{a priori} unknown endoscopic equipment settings.

The original structure of the dataset is as follows:
\begin{itemize}
    \item Kvasir Dataset Folds v2
    \begin{itemize}
        \item 1
        \begin{itemize}
            \item train
            \begin{itemize}
                \item dyed-lifted-polyps
                \item dyed-resection-margins
                \item esophagitis
                \item normal-cecum
                \item normal-pylorus
                \item normal-z-line
                \item polyps
                \item ulcerative-colitis
            \end{itemize}
            \item test
        \end{itemize}
        \item 2,3,4,5,6,7,8,9,10
    \end{itemize}
\end{itemize}
Each of the 10 folds consists of test and train sets containing 8 classes dyed-lifted-polyps, dyed-resection-margins,
esophagitis, normal-cecum, normal-pylorus, normal-z-line, polyps and ulcerative-colitis.
The images have different resolutions from 720x576 up to
1920x1072 pixels. 
Some samples in the dataset have large green annotations illustrating the position and configuration of the endoscope inside the bowel, by use of an electromagnetic imaging system (ScopeGuide, Olympus Europe).

We removed 30,430 images that contain green annotations. 
The distribution of the remaining images in the dataset with respect to resolution is as follows:
\begin{itemize}
    \item Resulting dataset without annotated images (total 49,570 images):
    \begin{itemize}
        \item 1920x1072 : 7.340
        \item 720x576 : 20.940
        \item 1280x1024 : 21.220
        \item 1350x1064 : 30
        \item 1350x1080 : 30
        \item 576x576 : 10
    \end{itemize}
    \item Resulting train split without annotated images (total 44,613 images):
    \begin{itemize}
        \item 1920x1072 : 6.606
        \item 720x576 : 18.846
        \item 1280x1024 : 19.098
        \item 1350x1064 : 27
        \item 1350x1080 : 27
        \item 576x576 : 9
    \end{itemize}
    \item Resulting train split without annotated images (total 4,957 images):
    \begin{itemize}
        \item 1920x1072 : 734
        \item 720x576 : 2.094
        \item 1280x1024 : 2.122
        \item 1350x1064 : 3
        \item 1350x1080 : 3
        \item 576x576 : 1
    \end{itemize}
\end{itemize}
In the resulting dataset, we choose 1280x1024 resolution set which has the highest number of images. 
The remaining dataset consists of 21,220 images in 1280x1024
resolution (19,098 images from train split and 2,122 images from test split), belonging to the classes esophagitis, normal-pylorus, normal-z-line, polyps and ulcerative-colitis. 
The classes such as dyed-lifted-polyps, dyed-resection-margins and normal-cecum contain low resolution images, which are not sutiable for super-resolution task.

The image distribution with respect to the classes is as follows:
\begin{itemize}
    \item Total number of images in each class of the remaining dataset (total 21,220 images):
    \begin{itemize}
        \item test\_dyed-lifted-polyps : 0
        \item test\_dyed-resection-margins : 0
        \item test\_esophagitis : 7.280
        \item test\_normal-cecum : 0
        \item test\_normal-pylorus : 7.300
        \item test\_normal-z-line : 6.490
        \item test\_polyps : 100
        \item test\_ulcerative-colitis : 50
    \end{itemize}
    \item Number of images in each class of the train split (total 19.098 images):
    \begin{itemize}
        \item train\_dyed-lifted-polyps : 0
        \item train\_dyed-resection-margins : 0
        \item train\_esophagitis : 6.552
        \item train\_normal-cecum : 0
        \item train\_normal-pylorus : 6.570
        \item train\_normal-z-line : 5.841
        \item train\_polyps : 90
        \item train\_ulcerative-colitis : 45
    \end{itemize}
    \item Number of images in each class of the test split (total 2,122 images):
    \begin{itemize}
        \item test\_dyed-lifted-polyps : 0
        \item test\_dyed-resection-margins : 0
        \item test\_esophagitis : 728
        \item test\_normal-cecum : 0
        \item test\_normal-pylorus : 730
        \item test\_normal-z-line : 649
        \item test\_polyps : 10
        \item test\_ulcerative-colitis : 5
    \end{itemize}
\end{itemize}

\section*{Appendix D: Metrics Detailed Results}
In this section; C1, C2, C3, C4 and C5 represents the esophagitis, normal-pylorus, normal-z-line, polyps and ulcerative colitis classes in Kvasir Dataset, respectively.
\input{tables/metrics_psnr}
\input{tables/metrics_ssim}
\input{tables/metrics_lpips}
\input{tables/metrics_gmsd}
\input{tables/detailed_mos_results}
\newpage

\newpage
\section*{Appendix F: Hyper-Parameter Analysis of EndoL2H Loss Function}
\input{tables/loss_params}

\newpage
\section*{Appendix G: Run Time Evaluation}
\label{app:runtime}
{
We present the runtime comparisons of the analysed deep SR methods: EndoL2H, DBPN, RCAN and SRGAN.  Runtimes for testing are computed using python function \textit{timeit} that encapsulates \textit{forward} function. Runtime of each network is evaluated using  NVIDIA® Tesla® V100 GPU.  $8\times$, $10\times$, and $12\times$ upscalings were performed. Table \ref{tab:run_time} demonstrates average runtime results. 
\input{tables/run_time}
}

\end{document}

%% file: tables/metrics_summary.tex
\newcolumntype{s}{>{\hsize=0.5\hsize}X}
\begin{table*}
\color{blue}
\centering
  \caption{\color{blue}SR performances of EndoL2H compared with other approaches in terms of PSNR, SSIM, LPIPS and GMSD. It can be seen that with the highest PSNR and SSIM values in all scales, our method proves its pixel-wise similarity and structural composition quality. Moreover, with the lowest LPIPS and GMSD scores among all compared approaches our proposed method show its highest perceptual features similarity.}
\begin{tabularx}{\textwidth}{|s|s|X|X|X|X|X|X|X|}
 \hline
\textbf{Class}				& \textbf{Scale}	& \textbf{EndoL2H}	& \textbf{EndoL2H-w/o-C}	& \textbf{EndoL2H-w/o-T}	& \textbf{DBPN}		& \textbf{RCAN}		& \textbf{SRGAN}	& \textbf{Bicubic}		\\ \hline
\multirow{3}{*}{PSNR $\uparrow$}		& 8x				& $\bold{36.10 \pm 0.28}$	& $31.68 \pm 0.61$	& $29.93 \pm 0.41$	& $33.85 \pm 0.44$	& $32.46 \pm 0.51$	& $29.48 \pm 0.47$	& $22.68 \pm 0.43$	\\ 
							& 10x				& $\bold{33.89 \pm 0.32}$	& $30.15 \pm 0.48$	& $28.64 \pm 0.31$	& $32.05 \pm 0.58$	& $30.87 \pm 0.60$	& $27.80 \pm 0.73$	& $20.61 \pm 0.31$	\\ 
							& 12x				& $\bold{32.55 \pm 0.45}$	& $28.32 \pm 0.71$	& $26.59 \pm 0.64$	& $30.24 \pm 0.59$	& $29.05 \pm 0.47$	& $26.11 \pm 0.34$	& $18.52 \pm 0.46$	\\ \cline{1-9}
\multirow{3}{*}{SSIM $\uparrow$}		& 8x				& $\bold{0.89 \pm 0.15}$	& $0.79 \pm 0.12$	& $0.75 \pm 0.14$	& $0.84 \pm 0.24$	& $0.81 \pm 0.21$	& $0.77 \pm 0.31$	& $0.59 \pm 0.42$	\\ 
							& 10x				& $\bold{0.85 \pm 0.1}$	& $0.76 \pm 0.16$	& $0.73 \pm 0.17$	& $0.81 \pm 0.27$	& $0.78 \pm 0.24$	& $0.74 \pm 0.33$	& $0.56 \pm 0.45$	\\ 
							& 12x				& $\bold{0.83 \pm 0.22}$	& $0.73 \pm 0.18$	& $0.69 \pm 0.20$	& $0.78 \pm 0.30$	& $0.75 \pm 0.26$	& $0.71 \pm 0.35$	& $0.52 \pm 0.48$	\\ \cline{1-9}
\multirow{3}{*}{LPIPS $\downarrow$}		& 8x				& $\bold{0.14 \pm 0.07}$	& $0.21 \pm 0.04$	& $0.25 \pm 0.06$	& $0.17 \pm 0.08$	& $0.18 \pm 0.07$	& $0.23 \pm 0.10$	& $0.28 \pm 0.14$	\\ 
							& 10x				& $\bold{0.16 \pm 0.09}$	& $0.24 \pm 0.07$	& $0.29 \pm 0.07$	& $0.20 \pm 0.09$	& $0.21 \pm 0.08$	& $0.26 \pm 0.12$	& $0.31 \pm 0.17$	\\ 
							& 12x				& $\bold{0.20 \pm 0.11}$	& $0.27 \pm 0.09$	& $0.31 \pm 0.09$	& $0.23 \pm 0.11$	& $0.24 \pm 0.10$	& $0.29 \pm 0.13$	& $0.33 \pm 0.18$	\\ \cline{1-9}
\multirow{3}{*}{GMSD $\downarrow$}		& 8x				& $\bold{0.11 \pm 0.07}$	& $0.18 \pm 0.04$	& $0.23 \pm 0.06$	& $0.14 \pm 0.08$	& $0.15 \pm 0.07$	& $0.21 \pm 0.10$	& $0.29 \pm 0.14$	\\ 
							& 10x				& $\bold{0.13 \pm 0.09}$	& $0.21 \pm 0.07$	& $0.27 \pm 0.07$	& $0.17 \pm 0.19$	& $0.18 \pm 0.08$	& $0.24 \pm 0.12$	& $0.32 \pm 0.17$	\\ 
							& 12x				& $\bold{0.17 \pm 0.11}$	& $0.24 \pm 0.09$	& $0.29 \pm 0.09$	& $0.20 \pm 0.14$	& $0.21 \pm 0.10$	& $0.27 \pm 0.13$	& $0.34 \pm 0.18$	\\ \cline{1-9}
\end{tabularx}
 \label{tab:metrics_summary}%
\end{table*}

%% file: tables/mos_results.tex
\newcolumntype{b}{>{\hsize=1.3\hsize}X}
\newcolumntype{B}{>{\hsize=1.7\hsize}X}
\newcolumntype{W}{>{\centering\hsize=\dimexpr5\hsize+10\tabcolsep+4\arrayrulewidth\relax}X|}
\begin{table*}[]
\color{blue}
\centering
  \caption{\color{blue}MOS Results of EndoL2H compared with other approaches: 30 gastroenterologists performed tests by assigning an integral score (1 for bad quality, 5 for excellent quality) to 15 random output images from the test dataset. 3 comparison metrics are considered to subjectively quantify the sharpness, suitability for diagnosis and detail level of the reconstructed super-resolution images.}
\begin{tabularx}{\textwidth}{|l|B|X|X|b|l|B|X|X|b|l|B|X|X|b|l|}
 \hline
 & 
\multicolumn{5}{W}{\begin{tabular}[c]{@{}c@{}}\textbf{MOS Sharpness}\\ \textbf{Results}\end{tabular}} & 
\multicolumn{5}{W}{\begin{tabular}[c]{@{}c@{}}\textbf{MOS Suitability for Diagnosis}\\ \textbf{Results}\end{tabular}} & 
\multicolumn{5}{W}{\begin{tabular}[c]{@{}c@{}}\textbf{MOS Detail Level} \\ \textbf{Results}\end{tabular}} \\ \hline
\textbf{Image}	& \textbf{EndoL2H}	& \textbf{DBPN}	& \textbf{RCAN}	& \textbf{SRGAN}	& \textbf{BI}	& \textbf{EndoL2H}	& \textbf{DBPN}	& \textbf{RCAN}	& \textbf{SRGAN}	& \textbf{BI}	& \textbf{EndoL2H}	& \textbf{DBPN}	& \textbf{RCAN}	& \textbf{SRGAN}	& \textbf{BI}	\\ \hline
\textbf{Mean}	& 4.41				& \textbf{4.42}	& 4.26			& 4.12				& 3.80			& \textbf{4.69}		& 4.58			& 4.50			& 4.39				& 2.30			& \textbf{4.75}		& 4.58			& 4.43			& 4.13				& 2.29	\\ \hline
\textbf{Std.}	& 0.24				& 0.23			& 0.26			& 0.31				& 0.37			& 0.16				& 0.26			& 0.22			& 0.18				& 0.38			& 0.21				& 0.35			& 0.36			& 0.32				& 0.67	\\ \hline
\textbf{Max.}	& 4.82				& 4.90			& 4.88			& 4.68				& 4.35			& 4.89				& 4.90			& 4.87			& 4.66				& 2.93			& 4.89				& 4.88			& 4.87			& 4.82				& 3.79	\\ \hline
\textbf{Min.}	& 4.00				& 4.07			& 3.70			& 3.48				& 3.12			& 4.43				& 4.14			& 4.05			& 3.98				& 1.55			& 4.14				& 3.73			& 3.94			& 3.72				& 1.15	\\ \hline
\end{tabularx}
 \label{tab:mos_results}%
\end{table*}

%% file: tables/ablation_sab.tex
\begin{table}[!h]
\color{blue}
    \centering
    \caption{\color{blue} Ablation study for spatial attention block. SR performance with and without attention in terms of PSNR, SSIM, LPIPS and GMSD.}
    \label{tab:ablation_sab}
\begin{tabularx}{\columnwidth}{|s|s|X|X|}
 \hline
\textbf{Class}	& \textbf{Scale}	& \textbf{EndoL2H}	& {\textbf{EndoL2H-} \newline \textbf{No Attention}}\\ \hline
\multirow{3}{*}{PSNR $\uparrow$}		& 8x						& $36.10 \pm 0.28$				& $34.15 \pm 0.84$	\\ 
							& 10x						& $33.89 \pm 0.32$				& $31.28 \pm 0.71$	\\ 
							& 12x						& $32.55 \pm 0.45$				& $28.07 \pm 0.64$	\\ \cline{1-4}
\multirow{3}{*}{SSIM $\uparrow$}		& 8x						& $0.89 \pm 0.15$				& $0.86 \pm 0.28$	\\ 
							& 10x						& $0.85 \pm 0.19$				& $0.80 \pm 0.23$	\\ 
							& 12x						& $0.83 \pm 0.22$				& $0.76 \pm 0.37$	\\ \cline{1-4}
\multirow{3}{*}{LPIPS $\downarrow$}		& 8x						& $0.14 \pm 0.07$				& $0.16 \pm 0.07$	\\ 
							& 10x						& $0.16 \pm 0.09$				& $0.22 \pm 0.15$	\\ 
							& 12x						& $0.20 \pm 0.11$				& $0.26 \pm 0.21$	\\ \cline{1-4}
\multirow{3}{*}{GMSD $\downarrow$}		& 8x						& $0.11 \pm 0.07$				& $0.15 \pm 0.07$	\\ 
							& 10x						& $0.13 \pm 0.09$				& $0.19 \pm 0.10$	\\ 
							& 12x						& $0.17 \pm 0.11$				& $0.22 \pm 0.19$	\\ \cline{1-4}
\end{tabularx}
\end{table}

%% file: tables/metrics_psnr.tex
\newcolumntype{s}{>{\hsize=0.5\hsize}X}

\begin{table*}[!h]
\color{blue}
\centering
  \caption{\color{blue}SR performance in terms of PSNR similarity metric.}
\begin{tabularx}{\textwidth}{|s|s|X|X|X|X|X|X|X|}
 \hline
\textbf{Class}				& \textbf{Scale}	& \textbf{EndoL2H}	& \textbf{EndoL2H-w/o-C}	& \textbf{EndoL2H-w/o-T}	& \textbf{DBPN}		& \textbf{RCAN}		& \textbf{SRGAN}	& \textbf{Bicubic}		\\ \hline
\multirow{3}{*}{C1}			& 8x				& $35.94 \pm 2.58$	& $31.36 \pm 2.58$	& $30.04 \pm 2.62$	& $33.32 \pm 2.54$	& $32.06 \pm 2.64$	& $29.60 \pm 2.85$	& $22.41 \pm 3.75$	\\ 
							& 10x				& $34.03 \pm 2.71$	& $30.43 \pm 2.74$	& $28.65 \pm 2.80$	& $31.43 \pm 2.73$	& $31.71 \pm 2.69$	& $27.81 \pm 2.88$	& $20.25 \pm 3.83$	\\ 
							& 12x				& $33.00 \pm 2.77$	& $29.17 \pm 2.85$	& $26.44 \pm 2.88$	& $29.99 \pm 2.82$	& $29.78 \pm 2.71$	& $26.13 \pm 3.01$	& $18.15 \pm 3.88$	\\ \cline{1-9}
\multirow{3}{*}{C2}			& 8x				& $35.83 \pm 2.60$	& $30.88 \pm 2.49$	& $29.92 \pm 2.78$	& $33.40 \pm 2.63$	& $33.34 \pm 2.68$	& $30.05 \pm 2.85$	& $22.33 \pm 3.62$	\\ 
							& 10x				& $33.42 \pm 2.64$	& $29.48 \pm 2.66$	& $29.15 \pm 2.74$	& $32.96 \pm 2.68$	& $29.86 \pm 2.75$	& $26.91 \pm 2.85$	& $20.86 \pm 3.73$	\\ 
							& 12x				& $32.65 \pm 2.81$	& $27.63 \pm 2.76$	& $25.83 \pm 2.86$	& $30.91 \pm 2.77$	& $29.09 \pm 2.70$	& $25.60 \pm 2.88$	& $18.37 \pm 3.81$	\\ \cline{1-9}
\multirow{3}{*}{C3}			& 8x				& $36.36 \pm 2.55$	& $31.91 \pm 2.52$	& $29.79 \pm 2.68$	& $34.51 \pm 2.65$	& $32.06 \pm 2.64$	& $28.86 \pm 2.82$	& $23.23 \pm 3.70$	\\ 
							& 10x				& $34.38 \pm 2.71$	& $30.10 \pm 2.77$	& $28.50 \pm 2.70$	& $31.46 \pm 2.65$	& $31.00 \pm 2.70$	& $28.72 \pm 2.94$	& $20.99 \pm 3.81$	\\ 
							& 12x				& $31.82 \pm 2.65$	& $29.06 \pm 2.78$	& $26.02 \pm 2.86$	& $29.41 \pm 2.82$	& $29.07 \pm 2.82$	& $26.47 \pm 2.87$	& $18.16 \pm 3.90$	\\ \cline{1-9}
\multirow{3}{*}{C4}			& 8x				& $35.87 \pm 2.48$	& $31.51 \pm 2.68$	& $29.31 \pm 2.62$	& $33.97 \pm 2.52$	& $32.76 \pm 2.71$	& $29.88 \pm 2.78$	& $22.25 \pm 3.68$	\\ 
							& 10x				& $33.85 \pm 2.63$	& $30.87 \pm 2.63$	& $28.19 \pm 2.73$	& $31.98 \pm 2.72$	& $30.73 \pm 2.76$	& $28.49 \pm 2.87$	& $20.69 \pm 3.79$	\\ 
							& 12x				& $33.00 \pm 2.78$	& $28.32 \pm 2.68$	& $27.36 \pm 2.78$	& $29.98 \pm 2.68$	& $29.00 \pm 2.85$	& $25.89 \pm 2.98$	& $19.40 \pm 3.94$	\\ \cline{1-9}
\multirow{3}{*}{C5}			& 8x				& $36.50 \pm 2.51$	& $32.71 \pm 2.49$	& $30.58 \pm 2.68$	& $34.04 \pm 2.67$	& $32.09 \pm 2.53$	& $29.02 \pm 2.76$	& $23.19 \pm 3.65$	\\ 
							& 10x				& $33.75 \pm 2.69$	& $29.88 \pm 2.72$	& $28.70 \pm 2.85$	& $32.40 \pm 2.65$	& $31.06 \pm 2.62$	& $27.06 \pm 2.87$	& $20.25 \pm 3.88$	\\ 
							& 12x				& $32.29 \pm 2.71$	& $27.41 \pm 2.76$	& $27.31 \pm 2.85$	& $30.92 \pm 2.69$	& $28.31 \pm 2.81$	& $26.47 \pm 2.95$	& $18.53 \pm 3.82$	\\ \cline{1-9}
\multirow{3}{*}{Average}	& 8x				& $36.10 \pm 0.28$	& $31.68 \pm 0.61$	& $29.93 \pm 0.41$	& $33.85 \pm 0.44$	& $32.46 \pm 0.51$	& $29.48 \pm 0.47$	& $22.68 \pm 0.43$	\\ 
							& 10x				& $33.89 \pm 0.32$	& $30.15 \pm 0.48$	& $28.64 \pm 0.31$	& $32.05 \pm 0.58$	& $30.87 \pm 0.60$	& $27.80 \pm 0.73$	& $20.61 \pm 0.31$	\\ 
							& 12x				& $32.55 \pm 0.45$	& $28.32 \pm 0.71$	& $26.59 \pm 0.64$	& $30.24 \pm 0.59$	& $29.05 \pm 0.47$	& $26.11 \pm 0.34$	& $18.52 \pm 0.46$	\\ \cline{1-9}
\end{tabularx}
 \label{tab:metrics_psnr}%
\end{table*}

%% file: tables/metrics_ssim.tex
\newcolumntype{s}{>{\hsize=0.5\hsize}X}

\begin{table*}[h]
\color{blue}
\centering
  \caption{\color{blue}SR performance in terms of SSIM similarity metric.}
\begin{tabularx}{\textwidth}{|s|s|X|X|X|X|X|X|X|}
 \hline
\textbf{Class}				& \textbf{Scale}	& \textbf{EndoL2H}	& \textbf{EndoL2H-w/o-C}	& \textbf{EndoL2H-w/o-T}	& \textbf{DBPN}		& \textbf{RCAN}		& \textbf{SRGAN}	& \textbf{Bicubic}		\\ \hline
\multirow{3}{*}{C1}			& 8x				& $0.89 \pm 0.16$	& $0.78 \pm 0.12$	& $0.75 \pm 0.13$	& $0.83 \pm 0.23$	& $0.80 \pm 0.21$	& $0.77 \pm 0.31$	& $0.59 \pm 0.43$	\\ 
							& 10x				& $0.86 \pm 0.20$	& $0.77 \pm 0.16$	& $0.73 \pm 0.18$	& $0.80 \pm 0.28$	& $0.80 \pm 0.23$	& $0.74 \pm 0.33$	& $0.55 \pm 0.46$	\\ 
							& 12x				& $0.84 \pm 0.22$	& $0.75 \pm 0.20$	& $0.69 \pm 0.21$	& $0.77 \pm 0.31$	& $0.77 \pm 0.25$	& $0.71 \pm 0.37$	& $0.51 \pm 0.48$	\\ \cline{1-9}
\multirow{3}{*}{C2}			& 8x				& $0.89 \pm 0.16$	& $0.77 \pm 0.10$	& $0.75 \pm 0.16$	& $0.83 \pm 0.25$	& $0.83 \pm 0.22$	& $0.78 \pm 0.31$	& $0.59 \pm 0.40$	\\ 
							& 10x				& $0.84 \pm 0.18$	& $0.75 \pm 0.15$	& $0.74 \pm 0.16$	& $0.83 \pm 0.27$	& $0.76 \pm 0.25$	& $0.72 \pm 0.33$	& $0.56 \pm 0.44$	\\ 
							& 12x				& $0.83 \pm 0.23$	& $0.72 \pm 0.18$	& $0.67 \pm 0.20$	& $0.79 \pm 0.30$	& $0.75 \pm 0.25$	& $0.70 \pm 0.34$	& $0.52 \pm 0.46$	\\ \cline{1-9}
\multirow{3}{*}{C3}			& 8x				& $0.90 \pm 0.16$	& $0.79 \pm 0.11$	& $0.74 \pm 0.14$	& $0.85 \pm 0.25$	& $0.80 \pm 0.21$	& $0.75 \pm 0.31$	& $0.61 \pm 0.42$	\\ 
							& 10x				& $0.86 \pm 0.20$	& $0.76 \pm 0.17$	& $0.72 \pm 0.16$	& $0.80 \pm 0.26$	& $0.78 \pm 0.24$	& $0.76 \pm 0.34$	& $0.57 \pm 0.45$	\\ 
							& 12x				& $0.82 \pm 0.20$	& $0.75 \pm 0.19$	& $0.68 \pm 0.20$	& $0.76 \pm 0.31$	& $0.75 \pm 0.27$	& $0.72 \pm 0.34$	& $0.51 \pm 0.48$	\\ \cline{1-9}
\multirow{3}{*}{C4}			& 8x				& $0.89 \pm 0.14$	& $0.78 \pm 0.14$	& $0.73 \pm 0.13$	& $0.84 \pm 0.23$	& $0.82 \pm 0.23$	& $0.78 \pm 0.30$	& $0.59 \pm 0.42$	\\ 
							& 10x				& $0.85 \pm 0.18$	& $0.78 \pm 0.14$	& $0.72 \pm 0.16$	& $0.81 \pm 0.28$	& $0.78 \pm 0.25$	& $0.75 \pm 0.33$	& $0.56 \pm 0.45$	\\ 
							& 12x				& $0.84 \pm 0.22$	& $0.73 \pm 0.17$	& $0.71 \pm 0.19$	& $0.77 \pm 0.28$	& $0.75 \pm 0.28$	& $0.71 \pm 0.36$	& $0.54 \pm 0.49$	\\ \cline{1-9}
\multirow{3}{*}{C5}			& 8x				& $0.90 \pm 0.15$	& $0.81 \pm 0.10$	& $0.76 \pm 0.14$	& $0.84 \pm 0.25$	& $0.80 \pm 0.19$	& $0.76 \pm 0.30$	& $0.61 \pm 0.41$	\\ 
							& 10x				& $0.85 \pm 0.20$	& $0.76 \pm 0.16$	& $0.73 \pm 0.19$	& $0.82 \pm 0.26$	& $0.79 \pm 0.22$	& $0.72 \pm 0.33$	& $0.55 \pm 0.47$	\\ 
							& 12x				& $0.83 \pm 0.21$	& $0.71 \pm 0.18$	& $0.70 \pm 0.20$	& $0.79 \pm 0.28$	& $0.74 \pm 0.27$	& $0.72 \pm 0.36$	& $0.52 \pm 0.47$	\\ \cline{1-9}
\multirow{3}{*}{Average}	& 8x				& $0.89 \pm 0.15$	& $0.79 \pm 0.12$	& $0.75 \pm 0.14$	& $0.84 \pm 0.24$	& $0.81 \pm 0.21$	& $0.77 \pm 0.31$	& $0.59 \pm 0.42$	\\ 
							& 10x				& $0.85 \pm 0.19$	& $0.76 \pm 0.16$	& $0.73 \pm 0.17$	& $0.81 \pm 0.27$	& $0.78 \pm 0.24$	& $0.74 \pm 0.33$	& $0.56 \pm 0.45$	\\ 
							& 12x				& $0.83 \pm 0.22$	& $0.73 \pm 0.18$	& $0.69 \pm 0.20$	& $0.78 \pm 0.30$	& $0.75 \pm 0.26$	& $0.71 \pm 0.35$	& $0.52 \pm 0.48$	\\ \cline{1-9}
\end{tabularx}
 \label{tab:metrics_ssim}%
\end{table*}

%% file: tables/metrics_lpips.tex
\newcolumntype{s}{>{\hsize=0.5\hsize}X}

\begin{table*}[!h]
\color{blue}
\centering
  \caption{\color{blue}SR performance in terms of LPIPS distortion metric.}
\begin{tabularx}{\textwidth}{|s|s|X|X|X|X|X|X|X|}
 \hline
\textbf{Class}				& \textbf{Scale}	& \textbf{EndoL2H}	& \textbf{EndoL2H-w/o-C}	& \textbf{EndoL2H-w/o-T}	& \textbf{DBPN}		& \textbf{RCAN}		& \textbf{SRGAN}	& \textbf{Bicubic}		\\ \hline
\multirow{3}{*}{C1}			& 8x				& $0.14 \pm 0.07$	& $0.20 \pm 0.05$	& $0.25 \pm 0.05$	& $0.16 \pm 0.07$	& $0.17 \pm 0.07$	& $0.23 \pm 0.10$	& $0.28 \pm 0.15$	\\ 
							& 10x				& $0.17 \pm 0.10$	& $0.25 \pm 0.08$	& $0.29 \pm 0.08$	& $0.19 \pm 0.10$	& $0.23 \pm 0.08$	& $0.26 \pm 0.12$	& $0.30 \pm 0.17$	\\ 
							& 12x				& $0.21 \pm 0.11$	& $0.29 \pm 0.10$	& $0.31 \pm 0.10$	& $0.22 \pm 0.12$	& $0.26 \pm 0.09$	& $0.29 \pm 0.14$	& $0.32 \pm 0.18$	\\ \cline{1-9}
\multirow{3}{*}{C2}			& 8x				& $0.14 \pm 0.08$	& $0.19 \pm 0.03$	& $0.25 \pm 0.07$	& $0.16 \pm 0.08$	& $0.20 \pm 0.07$	& $0.24 \pm 0.10$	& $0.28 \pm 0.13$	\\ 
							& 10x				& $0.15 \pm 0.09$	& $0.23 \pm 0.06$	& $0.30 \pm 0.07$	& $0.22 \pm 0.09$	& $0.19 \pm 0.09$	& $0.24 \pm 0.11$	& $0.31 \pm 0.16$	\\ 
							& 12x				& $0.20 \pm 0.12$	& $0.26 \pm 0.09$	& $0.29 \pm 0.10$	& $0.24 \pm 0.11$	& $0.24 \pm 0.09$	& $0.28 \pm 0.12$	& $0.33 \pm 0.17$	\\ \cline{1-9}
\multirow{3}{*}{C3}			& 8x				& $0.15 \pm 0.07$	& $0.21 \pm 0.04$	& $0.24 \pm 0.06$	& $0.18 \pm 0.08$	& $0.17 \pm 0.07$	& $0.21 \pm 0.10$	& $0.30 \pm 0.14$	\\ 
							& 10x				& $0.17 \pm 0.10$	& $0.24 \pm 0.08$	& $0.28 \pm 0.07$	& $0.19 \pm 0.09$	& $0.21 \pm 0.08$	& $0.28 \pm 0.12$	& $0.32 \pm 0.17$	\\ 
							& 12x				& $0.19 \pm 0.10$	& $0.29 \pm 0.09$	& $0.30 \pm 0.10$	& $0.21 \pm 0.12$	& $0.24 \pm 0.11$	& $0.30 \pm 0.12$	& $0.32 \pm 0.19$	\\ \cline{1-9}
\multirow{3}{*}{C4}			& 8x				& $0.14 \pm 0.06$	& $0.20 \pm 0.06$	& $0.23 \pm 0.05$	& $0.17 \pm 0.06$	& $0.19 \pm 0.08$	& $0.24 \pm 0.09$	& $0.28 \pm 0.14$	\\ 
							& 10x				& $0.16 \pm 0.09$	& $0.26 \pm 0.06$	& $0.28 \pm 0.07$	& $0.20 \pm 0.10$	& $0.21 \pm 0.09$	& $0.27 \pm 0.11$	& $0.31 \pm 0.16$	\\ 
							& 12x				& $0.21 \pm 0.12$	& $0.27 \pm 0.07$	& $0.33 \pm 0.08$	& $0.22 \pm 0.10$	& $0.24 \pm 0.11$	& $0.29 \pm 0.14$	& $0.35 \pm 0.19$	\\ \cline{1-9}
\multirow{3}{*}{C5}			& 8x				& $0.15 \pm 0.06$	& $0.23 \pm 0.03$	& $0.26 \pm 0.06$	& $0.17 \pm 0.09$	& $0.17 \pm 0.05$	& $0.22 \pm 0.09$	& $0.30 \pm 0.14$	\\ 
							& 10x				& $0.16 \pm 0.10$	& $0.24 \pm 0.07$	& $0.29 \pm 0.09$	& $0.21 \pm 0.09$	& $0.22 \pm 0.07$	& $0.24 \pm 0.11$	& $0.30 \pm 0.18$	\\ 
							& 12x				& $0.20 \pm 0.11$	& $0.25 \pm 0.09$	& $0.32 \pm 0.09$	& $0.24 \pm 0.10$	& $0.23 \pm 0.11$	& $0.30 \pm 0.13$	& $0.33 \pm 0.18$	\\ \cline{1-9}
\multirow{3}{*}{Average}	& 8x				& $0.14 \pm 0.07$	& $0.21 \pm 0.04$	& $0.25 \pm 0.06$	& $0.17 \pm 0.08$	& $0.18 \pm 0.07$	& $0.23 \pm 0.10$	& $0.28 \pm 0.14$	\\ 
							& 10x				& $0.16 \pm 0.09$	& $0.24 \pm 0.07$	& $0.29 \pm 0.07$	& $0.20 \pm 0.09$	& $0.21 \pm 0.08$	& $0.26 \pm 0.12$	& $0.31 \pm 0.17$	\\ 
							& 12x				& $0.20 \pm 0.11$	& $0.27 \pm 0.09$	& $0.31 \pm 0.09$	& $0.23 \pm 0.11$	& $0.24 \pm 0.10$	& $0.29 \pm 0.13$	& $0.33 \pm 0.18$	\\ \cline{1-9}
\end{tabularx}
 \label{tab:metrics_lpips}%
\end{table*}

%% file: tables/metrics_gmsd.tex
\newcolumntype{s}{>{\hsize=0.5\hsize}X}

\begin{table*}[!h]
\color{blue}
\centering
  \caption{\color{blue}SR performance in terms of GMSD distortion metric.}
\begin{tabularx}{\textwidth}{|s|s|X|X|X|X|X|X|X|}
 \hline
\textbf{Class}				& \textbf{Scale}	& \textbf{EndoL2H}	& \textbf{EndoL2H-w/o-C}	& \textbf{EndoL2H-w/o-T}	& \textbf{DBPN}		& \textbf{RCAN}		& \textbf{SRGAN}	& \textbf{Bicubic}		\\ \hline
\multirow{3}{*}{C1}			& 8x				& $0.11 \pm 0.07$	& $0.17 \pm 0.05$	& $0.23 \pm 0.05$	& $0.13 \pm 0.07$	& $0.14 \pm 0.07$	& $0.21 \pm 0.10$	& $0.29 \pm 0.15$	\\ 
							& 10x				& $0.14 \pm 0.10$	& $0.22 \pm 0.08$	& $0.27 \pm 0.08$	& $0.16 \pm 0.10$	& $0.20 \pm 0.08$	& $0.24 \pm 0.12$	& $0.31 \pm 0.17$	\\ 
							& 12x				& $0.18 \pm 0.11$	& $0.26 \pm 0.10$	& $0.29 \pm 0.10$	& $0.19 \pm 0.12$	& $0.23 \pm 0.09$	& $0.27 \pm 0.14$	& $0.33 \pm 0.18$	\\ \cline{1-9}
\multirow{3}{*}{C2}			& 8x				& $0.11 \pm 0.08$	& $0.16 \pm 0.03$	& $0.23 \pm 0.07$	& $0.13 \pm 0.08$	& $0.17 \pm 0.07$	& $0.22 \pm 0.10$	& $0.29 \pm 0.13$	\\ 
							& 10x				& $0.12 \pm 0.09$	& $0.20 \pm 0.06$	& $0.28 \pm 0.07$	& $0.19 \pm 0.09$	& $0.16 \pm 0.09$	& $0.22 \pm 0.11$	& $0.32 \pm 0.16$	\\ 
							& 12x				& $0.17 \pm 0.12$	& $0.23 \pm 0.09$	& $0.27 \pm 0.10$	& $0.21 \pm 0.11$	& $0.21 \pm 0.09$	& $0.26 \pm 0.12$	& $0.34 \pm 0.17$	\\ \cline{1-9}
\multirow{3}{*}{C3}			& 8x				& $0.12 \pm 0.07$	& $0.18 \pm 0.04$	& $0.22 \pm 0.06$	& $0.15 \pm 0.08$	& $0.14 \pm 0.07$	& $0.19 \pm 0.10$	& $0.31 \pm 0.14$	\\ 
							& 10x				& $0.14 \pm 0.10$	& $0.21 \pm 0.08$	& $0.26 \pm 0.07$	& $0.16 \pm 0.09$	& $0.18 \pm 0.08$	& $0.26 \pm 0.12$	& $0.33 \pm 0.17$	\\ 
							& 12x				& $0.16 \pm 0.10$	& $0.26 \pm 0.09$	& $0.28 \pm 0.10$	& $0.18 \pm 0.12$	& $0.21 \pm 0.11$	& $0.28 \pm 0.12$	& $0.33 \pm 0.19$	\\ \cline{1-9}
\multirow{3}{*}{C4}			& 8x				& $0.11 \pm 0.06$	& $0.17 \pm 0.06$	& $0.21 \pm 0.05$	& $0.14 \pm 0.06$	& $0.16 \pm 0.08$	& $0.22 \pm 0.09$	& $0.29 \pm 0.14$	\\ 
							& 10x				& $0.13 \pm 0.09$	& $0.23 \pm 0.06$	& $0.26 \pm 0.07$	& $0.17 \pm 0.10$	& $0.18 \pm 0.09$	& $0.25 \pm 0.11$	& $0.32 \pm 0.16$	\\ 
							& 12x				& $0.18 \pm 0.12$	& $0.24 \pm 0.07$	& $0.31 \pm 0.08$	& $0.19 \pm 0.10$	& $0.21 \pm 0.11$	& $0.27 \pm 0.14$	& $0.36 \pm 0.19$	\\ \cline{1-9}
\multirow{3}{*}{C5}			& 8x				& $0.12 \pm 0.06$	& $0.20 \pm 0.03$	& $0.24 \pm 0.06$	& $0.14 \pm 0.09$	& $0.14 \pm 0.05$	& $0.20 \pm 0.09$	& $0.31 \pm 0.14$	\\ 
							& 10x				& $0.13 \pm 0.10$	& $0.21 \pm 0.07$	& $0.27 \pm 0.09$	& $0.18 \pm 0.09$	& $0.19 \pm 0.07$	& $0.22 \pm 0.11$	& $0.31 \pm 0.18$	\\ 
							& 12x				& $0.17 \pm 0.11$	& $0.22 \pm 0.09$	& $0.30 \pm 0.09$	& $0.21 \pm 0.10$	& $0.20 \pm 0.11$	& $0.28 \pm 0.13$	& $0.34 \pm 0.18$	\\ \cline{1-9}
\multirow{3}{*}{Average}	& 8x				& $0.11 \pm 0.07$	& $0.18 \pm 0.04$	& $0.23 \pm 0.06$	& $0.14 \pm 0.08$	& $0.15 \pm 0.07$	& $0.21 \pm 0.10$	& $0.29 \pm 0.14$	\\ 
							& 10x				& $0.13 \pm 0.09$	& $0.21 \pm 0.07$	& $0.27 \pm 0.07$	& $0.17 \pm 0.09$	& $0.18 \pm 0.08$	& $0.24 \pm 0.12$	& $0.32 \pm 0.17$	\\ 
							& 12x				& $0.17 \pm 0.11$	& $0.24 \pm 0.09$	& $0.29 \pm 0.09$	& $0.20 \pm 0.11$	& $0.21 \pm 0.10$	& $0.27 \pm 0.13$	& $0.34 \pm 0.18$	\\ \cline{1-9}
\end{tabularx}
 \label{tab:metrics_gmsd}%
\end{table*}

%% file: tables/detailed_mos_results.tex
\newcolumntype{B}{>{\hsize=2.5\hsize}X}

\begin{table*}[!h]
\color{blue}
\centering
  \caption{\color{blue}Detailed MOS Results}
\begin{tabularx}{\textwidth}{|B|X|X|X|X|X|X|X|X|X|X|X|X|X|X|X||X|X|X|X|}
\hline
\multicolumn{20}{|>{\centering\hsize=\dimexpr20\hsize+21\tabcolsep+19\arrayrulewidth\relax}X|}{\textbf{MOS SHARPNESS RESULTS}}                                                                                                      \\ \hline
\textbf{Image} & \textbf{\#1} & \textbf{\#2} & \textbf{\#3} & \textbf{\#4} & \textbf{\#5} & \textbf{\#6} & \textbf{\#7} & \textbf{\#8} & \textbf{\#9} & \textbf{\#10} & \textbf{\#11} & \textbf{\#12} & \textbf{\#13} & \textbf{\#14} & \textbf{\#15} & \textbf{Mean}  & \textbf{Std.}   & \textbf{Max.}   & \textbf{Min.} \\ \hline
\textbf{EndoL2H}                & 4.82 & 4.38 & 4.36 & 4.47 & 4.37 & 4.10 & 4.17 & 4.41 & 4.00 & 4.59 & 4.25 & 4.51 & 4.78 & 4.26 & 4.70 &  4.41 & 0.24 & 4.82 & 4.00 \\ \hline
\textbf{DBPN}        & 4.52 & 4.40 & 4.90 & 4.69 & 4.30 & 4.50 & 4.17 & 4.16 & 4.27 & 4.67 & 4.07 & 4.54 & 4.53 & 4.16 & 4.39 &  4.42 & 0.23 & 4.90 & 4.07  \\ \hline
\textbf{RCAN}        & 4.19 & 4.26 & 4.24 & 4.32 & 4.37 & 4.88 & 4.17 & 4.32 & 3.86 & 4.20 & 3.70 & 4.32 & 4.27 & 4.37 & 4.44 &  4.26 & 0.26 & 4.88 & 3.70  \\ \hline
\textbf{SRGAN}                  & 3.48 & 4.04 & 4.17 & 4.24 & 3.86 & 4.46 & 4.20 & 4.35 & 4.49 & 3.93 & 4.19 & 3.83 & 4.07 & 3.88 & 4.68 &  4.12 & 0.31 & 4.68 & 3.48  \\ \hline
\textbf{Bicubic} & 3.63 & 3.89 & 3.68 & 3.73 & 4.25 & 4.15 & 3.37 & 3.12 & 4.35 & 3.73 & 3.94 & 4.34 & 3.37 & 3.91 & 3.53 &  3.80 & 0.37 & 4.35 & 3.12  \\ \hline
\multicolumn{20}{|>{\centering\hsize=\dimexpr20\hsize+21\tabcolsep+19\arrayrulewidth\relax}X|}{\textbf{MOS SUITABILITY FOR DIAGNOSIS RESULTS}}                                                                                      \\ \hline
\textbf{Image} & \textbf{\#1} & \textbf{\#2} & \textbf{\#3} & \textbf{\#4} & \textbf{\#5} & \textbf{\#6} & \textbf{\#7} & \textbf{\#8} & \textbf{\#9} & \textbf{\#10} & \textbf{\#11} & \textbf{\#12} & \textbf{\#13} & \textbf{\#14} & \textbf{\#15} & \textbf{Mean}  & \textbf{Std.}   & \textbf{Max.}   & \textbf{Min.} \\ \hline
\textbf{EndoL2H}                & 4.43 & 4.57 & 4.43 & 4.83 & 4.74 & 4.87 & 4.77 & 4.58 & 4.59 & 4.57 & 4.85 & 4.78 & 4.89 & 4.85 & 4.61 &  4.69 & 0.16 & 4.89 & 4.43  \\ \hline
\textbf{DBPN}        & 4.54 & 4.46 & 4.62 & 4.84 & 4.27 & 4.81 & 4.88 & 4.83 & 4.80 & 4.57 & 4.36 & 4.14 & 4.17 & 4.90 & 4.54 &  4.58 & 0.26 & 4.90 & 4.14  \\ \hline
\textbf{RCAN}        & 4.57 & 4.56 & 4.76 & 4.51 & 4.71 & 4.16 & 4.39 & 4.47 & 4.87 & 4.33 & 4.49 & 4.71 & 4.42 & 4.05 & 4.47 &  4.50 & 0.22 & 4.87 & 4.05  \\ \hline
\textbf{SRGAN}                  & 4.10 & 4.51 & 4.41 & 4.44 & 4.47 & 4.66 & 4.35 & 3.98 & 4.54 & 4.23 & 4.35 & 4.37 & 4.40 & 4.61 & 4.39 &  4.39 & 0.18 & 4.66 & 3.98  \\ \hline
\textbf{Bicubic} & 1.77 & 2.54 & 2.39 & 1.83 & 2.38 & 2.43 & 2.93 & 2.33 & 2.37 & 2.13 & 2.54 & 1.95 & 1.55 & 2.73 & 2.64 &  2.30 & 0.38 & 2.93 & 1.55  \\ \hline
\multicolumn{20}{|>{\centering\hsize=\dimexpr20\hsize+21\tabcolsep+19\arrayrulewidth\relax}X|}{\textbf{MOS DETAIL LEVEL RESULTS}}                                                                                                   \\ \hline
\textbf{Image} & \textbf{\#1} & \textbf{\#2} & \textbf{\#3} & \textbf{\#4} & \textbf{\#5} & \textbf{\#6} & \textbf{\#7} & \textbf{\#8} & \textbf{\#9} & \textbf{\#10} & \textbf{\#11} & \textbf{\#12} & \textbf{\#13} & \textbf{\#14} & \textbf{\#15} & \textbf{Mean}  & \textbf{Std.}   & \textbf{Max.}   & \textbf{Min.} \\ \hline
\textbf{EndoL2H}               & 4.87 & 4.86 & 4.85 & 4.87 & 4.46 & 4.60 & 4.14 & 4.59 & 4.80 & 4.87 & 4.89 & 4.87 & 4.86 & 4.89 & 4.82 &  4.75 & 0.21 & 4.89 & 4.14  \\ \hline
\textbf{DBPN}               & 4.09 & 4.72 & 3.73 & 4.61 & 4.81 & 4.65 & 4.76 & 4.78 & 4.84 & 4.74 & 4.85 & 4.88 & 4.74 & 4.38 & 4.04 &  4.58 & 0.35 & 4.88 & 3.73  \\ \hline
\textbf{RCAN}               & 4.83 & 4.87 & 3.97 & 4.67 & 4.86 & 4.82 & 3.94 & 3.98 & 4.12 & 4.50 & 4.46 & 4.15 & 4.27 & 4.16 & 4.84 &  4.43 & 0.36 & 4.87 & 3.94  \\ \hline
\textbf{SRGAN}                 & 4.49 & 4.42 & 3.98 & 3.72 & 4.28 & 4.40 & 3.95 & 4.38 & 4.82 & 3.85 & 4.28 & 3.82 & 3.84 & 3.88 & 3.90 &  4.13 & 0.32 & 4.82 & 3.72  \\ \hline
\textbf{Bicubic} & 2.23 & 1.15 & 2.60 & 1.44 & 2.19 & 2.17 & 2.19 & 3.79 & 1.89 & 2.17 & 3.24 & 2.45 & 1.60 & 2.59 & 2.68 &  2.29 & 0.67 & 3.79 & 1.15  \\ \hline
\end{tabularx}
 \label{tab:detailed_mos}%
\end{table*}

%% file: tables/loss_params.tex
\begin{table}[!h]
\color{blue}
    \caption{\color{blue} \textbf{Loss hyper-parameters.} We investigate the performance of the network for different EndoL2H-loss parameter settings. Parameter setting $\alpha=0.35$, $\beta=0.20$ and $\gamma=0.15$ show best performance in terms of PSNR, SSIM, LPIPS and GMSD metrics.}
    \begin{tabularx}{\columnwidth}{|X|X|X||X|X|X|X|}
\hline 
\multicolumn{3}{|>{\centering\hsize=\dimexpr3\hsize+6\tabcolsep+2\arrayrulewidth\relax}X||}{\textbf{Parameters of EndoL2H Loss}} & 
\multicolumn{4}{>{\centering\hsize=\dimexpr4\hsize+8\tabcolsep+3\arrayrulewidth\relax}X|}{\textbf{Training Performance}} \\ \hline
$\mathbf{\alpha}$	& $\mathbf{\beta}$	& $\mathbf{\gamma}$	& \textbf{PSNR}	& \textbf{SSIM}	& \textbf{LPIPS}	& \textbf{GMSD}	\\ \hline
0.25			& 0.25			& 0.25			& 30.45				& 0.75			& 0.18			& 0.15	\\ \hline
0.15			& 0.40			& 0.35			& 31.28				& 0.57			& 0.21			& 0.19	\\ \hline
\textbf{0.35}	& \textbf{0.20}	& \textbf{0.15}	& \textbf{36.19}	& \textbf{0.88}	& \textbf{0.15}	& \textbf{0.12}	\\ \hline
0.05			& 0.55			& 0.40			& 25.47				& 0.46			& 0.29			& 0.24	\\ \hline
0.15			& 0.15			& 0.65			& 33.49				& 0.65			& 0.21			& 0.20	\\ \hline
0.70			& 0.20			& 0.05			& 34.69				& 0.60			& 0.23			& 0.24	\\ \hline
0.50			& 0.30			& 0.30			& 32.47				& 0.64			& 0.20			& 0.16	\\ \hline
0.05			& 0.10			& 0.75			& 26.17				& 0.45			& 0.31			& 0.26	\\ \hline
0.15			& 0.70			& 0.05			& 28.34				& 0.53			& 0.26			& 0.23	\\ \hline
0.45			& 0.05			& 0.35			& 32.18				& 0.68			& 0.19			& 0.20	\\ \hline
    \end{tabularx}
    \label{tab:loss_params}%
\end{table}

%% file: tables/run_time.tex
\begin{table}[!h]
\color{blue}
    \centering
    \caption{\color{blue}Runtime evaluations for 8x, 10x, 12x super-resolution.}
    \label{tab:run_time}
\begin{tabular}{|l|l|l|l|}
 \hline
\textbf{Class}	& \textbf{8x}	& \textbf{10x}	& \textbf{12x}	\\ \hline
EndoL2H			& $116.21$		& $130.56$		& $147.63$		\\ \hline
DBPN			& $98.15$		& $112.10$		& $129.49$		\\ \hline
RCAN			& $89.79$		& $101.06$		& $121.39$		\\ \hline
SRGAN			& $80.14$		& $90.58$		& $106.61$		\\ \hline
\end{tabular}
\end{table}